\documentclass{article}

\PassOptionsToPackage{numbers, sort&compress}{natbib}
\usepackage[main, final]{neurips_2026}
\usepackage{amsmath}
\usepackage{amssymb}
\usepackage{mathtools}
\usepackage{amsthm}
\usepackage{multirow}
\usepackage{graphicx}
\usepackage{subcaption}

\usepackage{tablefootnote}
\usepackage{etoc}
\makeatletter
\let\origaddcontentsline\addcontentsline
\makeatother

\usepackage{url}

\usepackage{mathtools}
\usepackage{adjustbox}

\usepackage{setspace}
\usepackage{algorithm}
\usepackage{eqnarray}

\usepackage{thm-restate} 

\usepackage[utf8]{inputenc} 
\usepackage[T1]{fontenc}    

\usepackage{booktabs}       

\usepackage{amsfonts}       
\usepackage{nicefrac}       
\usepackage{microtype}      
\usepackage{xcolor}         
\usepackage{subfiles}       
\usepackage{standalone}     

\usepackage{hyperref} 
\usepackage[capitalize,noabbrev]{cleveref}

\theoremstyle{plain}
\newtheorem{theorem}{Theorem}[section]
\newtheorem{proposition}[theorem]{Proposition}

\theoremstyle{definition}
\newtheorem{definition}[theorem]{Definition}

\theoremstyle{remark}

\newcommand{\norm}[1]{\lVert {#1} \rVert} 

\newcommand{\VINE}{VINE\xspace}

\newcommand{\ignore}[1]{}

\usepackage{dsfont}		

\usepackage{acronym}		

\usepackage[labelfont={bf,small},labelsep=colon,font=small]{caption}	
\usepackage{pifont}

\usepackage{tikz}		
\usetikzlibrary{calc,patterns,positioning}

\usepackage{booktabs}		
\usepackage[inline,shortlabels]{enumitem}		
\setenumerate{itemsep=0pt,topsep=1pt,left=0pt}
\setitemize{itemsep=0pt,topsep=1pt,left=0pt}

\usepackage{xspace}		

\usepackage{thmtools}		
\usepackage{thm-restate}		

\theoremstyle{plain}

\newtheorem*{corollary*}{Corollary}		

\newcommand{\debug}[1]{#1}		

\newcommand{\newmacro}[2]{\newcommand{#1}{\debug{#2}}}		
\newcommand{\newop}[2]{\DeclareMathOperator{#1}{\debug{#2}}}		

\DeclarePairedDelimiterX{\setdef}[2]{\{}{\}}{#1:#2}		
\DeclarePairedDelimiterXPP{\exclude}[1]{\mathopen{}\setminus}{\{}{\}}{}{#1}

\DeclareMathOperator{\bigoh}{\mathcal{O}}		
\DeclarePairedDelimiterXPP{\bigof}[1]{\bigoh}{(}{)}{}{#1}		
\DeclareMathOperator{\dist}{dist}		
\DeclareMathOperator{\relint}{ri}		

\newcommand{\alt}[1]{#1'}		
\newcommand{\altalt}[1]{#1''}		

\newmacro{\dd}{\:d}		

\newmacro{\const}{C}		
\newmacro{\constalt}{B}		
\newmacro{\constdual}{\const_{\texttt{dual}}}		
\newmacro{\conststoch}{\constalt}		

\newmacro{\multi}{\texttt{H}}		

\newmacro{\coef}{\lambda}		
\newmacro{\param}{\theta}		
\newmacro{\params}{\Theta}		

\newmacro{\pexp}{r}		
\newmacro{\qexp}{q}		
\newmacro{\rexp}{r}		

\newmacro{\beforestart}{0}		
\newmacro{\start}{1}		
\newmacro{\afterstart}{2}		
\newmacro{\running}{\start,\afterstart,\dotsc\,}		

\newmacro{\run}{t}		
\newmacro{\runprev}{\run-1}		
\newmacro{\runalt}{s}		
\newmacro{\runaltalt}{\alt \run}		
\newmacro{\nRuns}{T}		
\newmacro{\runs}{\mathcal{\nRuns}}		

\newmacro{\state}{X}		
\newmacro{\stateavg}{\bar\state}		
\newmacro{\stateopt}{\tilde\state}		
\newmacro{\statealt}{\alt\state}		
\newmacro{\query}{x}		
\newmacro{\out}{\bar\query}		

\newop{\Eq}{Eq}		
\newop{\Nash}{NE}		

\newcommand{\gap}{\Delta}
\newop{\brep}{br}		
\newop{\regstoch}{\tilde{\reg}}		
\newop{\preg}{Reg}		

\newop{\val}{val}		
\newmacro{\play}{i}		
\newmacro{\playalt}{j}		
\newmacro{\playaltalt}{k}		
\newmacro{\nPlayers}{N}		
\newmacro{\players}{\mathcal{\nPlayers}}		

\newmacro{\pure}{a}		
\newmacro{\purealt}{\beta}		
\newmacro{\purealtalt}{\gamma}		
\newmacro{\nPures}{n}		
\newmacro{\pures}{\mathcal{A}}		

\newmacro{\strat}{p}		
\newmacro{\stratalt}{\alt\strat}		
\newmacro{\strataltalt}{\altalt\strat}		
\newmacro{\strats}{\mathcal{X}}		
\newmacro{\intstrats}{\strats^{\circle}}		

\newmacro{\pay}{u}		
\newmacro{\payv}{v}		
\newmacro{\loss}{\ell}
\newmacro{\lossv}{\sample}
\newmacro{\pot}{\Phi}		
\newmacro{\meanpot}{\pot}		

\newmacro{\game}{\Gamma}		
\newmacro{\meangame}{\game}		
\newmacro{\gameall}{\game(\players,\points,\loss)}		

\newmacro{\fingame}{\Gamma}		
\newmacro{\fingameall}{\Gamma(\players,\pures,\pay)}		

\newmacro{\gmat}{g}		
\newmacro{\gdist}{\dist_{\gmat}}
\newmacro{\mfld}{M}		
\newmacro{\form}{\omega}		

\newmacro{\tvec}{z}		
\newmacro{\uvec}{u}		

\newmacro{\ball}{\mathbb{B}}		

\newmacro{\vecspace}{\mathcal{V}}		
\newmacro{\subspace}{\mathcal{W}}		

\newmacro{\bvec}{e}		
\newmacro{\bvecs}{\mathcal{E}}		

\newmacro{\coord}{i}		
\newmacro{\coordalt}{\coord^{\prime}}		
\newmacro{\coordaltalt}{\coordalt^{\prime}}		
\newmacro{\nCoords}{d}		
\newmacro{\dims}{\nCoords}		
\newmacro{\vdim}{\nCoords}		

\newmacro{\pspace}{\vecspace}		
\newmacro{\dspace}{\vecspace^{\ast}}		

\newmacro{\pstate}{z}		

\newmacro{\dstate}{Y}		
\newmacro{\dvec}{{v}}		
\newmacro{\disdstate}{\hat\dstate}		

\newmacro{\Anc}{\textsf{r}}
\newmacro{\anchor}{R}
\newmacro{\ptest}{\tilde{\pstate}}		
\newmacro{\dtest}{\tilde{\drecom}}		
\newmacro{\testsignal}{\tilde{\signal}}		
\newmacro{\precom}{\pstate}	
\newmacro{\drecom}{\dstate}		

\newmacro{\dispstate}{Z}		
\newmacro{\dispstatealt}{\bar\dispstate}		

\newmacro{\dpoint}{y}		

\newmacro{\dpointave}{\bar\dpoint}		
\newmacro{\dpointalt}{W}		
\newmacro{\dpointaltalt}{\altalt\dpoint}		
\newmacro{\dpoints}{\mathcal{Y}}		

\newmacro{\sgrad}{g}
\newmacro{\shess}{H}
\newmacro{\vargrad}{\sigma_g}
\newmacro{\varhess}{\sigma_H}

\newmacro{\wgrad}{a}                
\newmacro{\wgradsum}{A}
\newmacro{\wavg}{b}                 
\newmacro{\wavgsum}{B}

\newmacro{\eucdiam}{D}

\newmacro{\mat}{\mathbf{M}}		
\newmacro{\hmat}{\mathbf{H}}		

\newmacro{\ones}{\mathbf{1}}		
\newmacro{\eye}{\mathbf{I}}		
\newmacro{\zer}{\mathbf{0}}		

\DeclarePairedDelimiterXPP{\dnormdef}[1]{}{\lVert}{\rVert}{_{\ast}}{#1}
\DeclarePairedDelimiterXPP{\dnorm}[1]{}{\lVert}{\rVert}{_{\ast}}{#1}

\DeclarePairedDelimiterXPP{\onenorm}[1]{}{\lVert}{\rVert}{_{1}}{#1}		
\DeclarePairedDelimiterXPP{\twonorm}[1]{}{\lVert}{\rVert}{_{2}}{#1}		
\DeclarePairedDelimiterXPP{\supnorm}[1]{}{\lVert}{\rVert}{_{\infty}}{#1}		

\DeclarePairedDelimiterXPP{\altdnorm}[1]{}{\lVert}{\rVert'}{_{\ast}}{#1}

\DeclarePairedDelimiterX{\braket}[2]{\langle}{\rangle}{#1\mathopen{},\mathopen{}#2}

\DeclarePairedDelimiterX{\inner}[2]{\langle}{\rangle}{#1,#2}		
\newmacro{\cartprod}{\bigtimes}

\newop{\Opt}{Opt}		
\newop{\Sol}{Sol}		
\newop{\orcl}{\mathsf{G}}		

\newmacro{\obj}{f}		
\newmacro{\objalt}{\alt \obj}		
\newmacro{\sobj}{F}		
\newmacro{\func}{\textsl{g}}

\newmacro{\gvec}{g}		
\newmacro{\oper}{A}		
\newmacro{\vecfield}{v}		
\newmacro{\seed}{\sample}		

\newmacro{\gbound}{G}		
\newmacro{\lips}{L}

\newmacro{\strong}{\kappa}		
\newmacro{\smooth}{\lips}		

\newmacro{\cvx}{\mathcal{K}}		
\newmacro{\compact}{\mathcal{X}}		
\newmacro{\subd}{\partial}		
\newmacro{\subsel}{\nabla}		

\newmacro{\minmax}{L}		

\newmacro{\minvar}{\theta}		
\newmacro{\minvaralt}{\alt\minvar}		
\newmacro{\minvars}{\Theta}		

\newmacro{\maxvar}{\phi}		
\newmacro{\maxvaralt}{\alt\maxvar}		
\newmacro{\maxvars}{\Phi}		

\newmacro{\hreg}{h}		
\newmacro{\proxdom}{\points_{\hreg}}		

\newmacro{\breg}{D}		
\newmacro{\mprox}{P}		

\newmacro{\fench}{F}		
\newmacro{\mirror}{Q}		

\newmacro{\hstr}{K_{\hreg}}		
\newmacro{\bregdiam}{B_{\hreg}}		
\newmacro{\range}{R_{\hreg}}		

\DeclarePairedDelimiterXPP{\proxof}[2]{\mprox_{#1}}{(}{)}{}{#2}		
\DeclarePairedDelimiterXPP{\bregof}[2]{\breg}{(}{)}{}{#1,#2}		
\DeclarePairedDelimiterXPP{\fenchof}[2]{\fench}{(}{)}{}{#1,#2}		

\newmacro{\zone}{\mathbb{D}}		

\newop{\Eucl}{\Pi}		
\newop{\logit}{\Lambda}		
\newmacro{\subgrad}{W}

\newmacro{\point}{x}		
\newmacro{\pointalt}{\alt\point}		
\newmacro{\pointaltalt}{\altalt\point}		
\newmacro{\points}{\mathcal{X}}		
\newmacro{\intpoints}{\relint\points}		

\newmacro{\base}{\point^{\ast}}		
\newmacro{\basealt}{u^{\ast}}		

\newmacro{\real}{x}
\newmacro{\realalt}{\alt\real}
\newmacro{\realaltalt}{\altalt \real}

\newmacro{\open}{\mathcal{U}}		
\newmacro{\closed}{\mathcal{C}}		
\newmacro{\cpt}{\mathcal{K}}		
\newmacro{\nhd}{\mathcal{U}}		

\newop{\ex}{\mathbb{E}}		
\newop{\prob}{\mathbb{P}}		
\newop{\simplex}{\Delta}		

\DeclarePairedDelimiterXPP{\exof}[1]{\ex}{[}{]}{}{
	 #1}

\DeclarePairedDelimiterXPP{\probof}[1]{\prob}{(}{)}{}{
	 #1}

\newmacro{\sample}{\omega}		
\newmacro{\samples}{\Omega}		

\newmacro{\filter}{\mathcal{F}}		
\newmacro{\probspace}{(\samples,\filter,\prob)}		

\newmacro{\event}{E}       
\newmacro{\eventalt}{H}       
\newmacro{\mean}{\mu}		
\newmacro{\sdev}{\sigma}		
\newmacro{\variance}{\sdev^{2}}		

\newmacro{\step}{\gamma}    
\newmacro{\stepalt}{\gamma}		
\newmacro{\stepaltalt}{\lambda}		
\newmacro{\stepada}{\theta}		
\newmacro{\stepscale}{\beta_0}

\newmacro{\learn}{\eta}		
\newmacro{\dstep}{\psi}		

\newmacro{\proper}{\tau}		

\newmacro{\signal}{\gvec}		
\newmacro{\altsignal}{\bar\signal}		
\newmacro{\error}{Z}		
\newmacro{\noise}{U}		
\newmacro{\bias}{b}		
\newmacro{\brown}{W}		

\newmacro{\serror}{\theta}		
\newmacro{\snoise}{\xi}		
\newmacro{\sbias}{\psi}		

\newmacro{\sbound}{M}		
\newmacro{\bbound}{B}		
\newmacro{\noisepar}{\sdev}		
\newmacro{\noisevar}{\texttt{var}}		

\newmacro{\cost}{c}

\newmacro{\boundcoord}{a}  
\newmacro{\smoothcoord}{b}		

\newmacro{\noiseave}{\tilde{\noise}}  
\newmacro{\noisetest}{\tilde{\noise}}  

\newmacro{\mindiff}{\rho}
\newmacro{\diff}{H}

\newmacro{\resource}{s}
\newmacro{\nResources}{d}
\newmacro{\resources}{\mathcal{S}}
\newmacro{\inflow}{\rho}
\newmacro{\load}{x}

\newmacro{\spectron}{\mathcal{D}}
\newmacro{\nInput}{M}
\newmacro{\nOutput}{N}
\newmacro{\chanmat}{\mathbf{H}}
\newmacro{\covmat}{\mathbf{X}}
\newmacro{\bx}{\mathbf{x}}
\newmacro{\by}{\mathbf{y}}
\newmacro{\bz}{\mathbf{z}}
\newmacro{\capa}{R}
\newmacro{\power}{P}

\newmacro{\maxsel}{m}

\newmacro{\shape}{\chi}
\newmacro{\diampoints}{\norm{\points}}

\newmacro{\underconst}{\const}
\newmacro{\slow}{\mathrm{slow}}
\newmacro{\fast}{\mathrm{fast}}

\newmacro{\numlvl}{\alpha}
\newmacro{\mtypeset}{\mathbb{S}}
\newmacro{\numseq}{M}

\def\1{\bm{1}}

\DeclareMathAlphabet{\mathsfit}{\encodingdefault}{\sfdefault}{m}{sl}
\SetMathAlphabet{\mathsfit}{bold}{\encodingdefault}{\sfdefault}{bx}{n}

\newcommand{\E}{\mathbb{E}}

\newcommand{\R}{\mathbb{R}}

\newcommand{\reg}{\lambda}

\DeclareMathOperator{\sign}{sign}

\def\R{\mathbb{R}}

\def \R{\mathbb{R}}

\usepackage{algorithm}
\usepackage{algorithmic}

\usepackage{tikz}
\usetikzlibrary{automata, positioning, arrows, quotes, shapes}
\tikzset{every node/.style={ellipse, font=\footnotesize, align=center}}
\tikzset{
    node distance = 15mm and 15mm,
    V/.style = {draw, fill=white},
    every edge quotes/.style = {auto, font=\scriptsize, sloped, above}
}

\newcommand{\threefiguresstacked}[3]{
\begin{tikzpicture}
  \def\w{4cm}
  \def\gap{0.8cm}

  \node (A) at (0,0) {#1};

  \node (B) at (0,-\gap)
    [below of=A]
    {#2};

  \node (C) at (0,-2*\gap)
    [below of=B]
    {#3};
\end{tikzpicture}
}

\newcommand{\threefiguresgrid}[3]{
\begin{tikzpicture}
  \def\sep{2.8cm}
  \def\vsep{2cm}

  \node (A) at (-\sep/1.8,0) {#1};

  \node (B) at (\sep/1.8,0) {#2};

  \node (C) at (0,-\vsep) {#3};
\end{tikzpicture}
}

\usepackage{wrapfig}
\usepackage{lipsum}

\usepackage{subcaption}

\title{Non-Parametric Mutual Information Estimation with Vine Copulas}

\author{%
  Sigurd Holmsen\textsuperscript{1,3}\thanks{Equal contribution} \\
  \texttt{sigurho@ifi.uio.no} \\
  \And
  Berit Omli Øksnes\textsuperscript{2,3}\footnotemark[1] \\
  \texttt{beritoo@math.uio.no} \\
  \AND
  Ingrid Hobæk Haff\textsuperscript{2,3,5} \\
  \texttt{ingrihaf@math.uio.no} \\
  \And
  Sylvia Richardson\textsuperscript{3,4} \\
  \texttt{sylvia.richardson@} \\
  \texttt{mrc-bsu.cam.ac.uk} \\
  \And
  Ali Ramezani-Kebrya\textsuperscript{1,3,5} \\
  \texttt{ali@ifi.uio.no} \\
  \AND
  \normalfont\small
  \textsuperscript{1}Department of Informatics, University of Oslo \\
  \normalfont\small
  \textsuperscript{2}Department of Mathematics, University of Oslo \\
  \normalfont\small
  \textsuperscript{3}Integreat -- Norwegian Centre for Knowledge-driven Machine Learning \\
  \normalfont\small
  \textsuperscript{4}MRC Biostatistics Unit, University of Cambridge \\
  \normalfont\small
  \textsuperscript{5}TRUST -- The Norwegian Centre for Trustworthy AI 
}

\begin{document}

\maketitle

\begin{abstract}
    We introduce \VINE: a non-parametric mutual information (MI) estimator based on simplified D-vine copulas. Approximating the joint density with a simplified D-vine allows the density ratio to simplify to a product of pair copulas, which we can estimate non-parametrically. This structural simplification yields \VINE, a flexible and computationally efficient MI estimator. We compare \VINE against SotA neural MI estimators on a range of datasets, both under settings where the simplified D-vine assumption holds and where it is violated, demonstrating competitive accuracy at a lower computational cost for low- to medium-dimensional datasets.
\end{abstract}

\section{Introduction}
\label{sec:introduction}
Mutual information (MI) is an information-theoretic measure that quantifies statistical dependence between two random vectors by measuring the reduction in uncertainty about one given the other \citep{shannon1948mathematical}.
MI estimation is the key challenge in important applications, e.g., channel capacity in digital communication~\citep{shannon1948mathematical},
gene regulatory networks in biology \citep{margolin2006aracne}, and feature selection in data science \cite{peng2005feature}. In machine learning with neural networks, efficient MI estimation paves the way for achieving representations of an input that are sufficient for prediction of the ground-truth output while carrying minimal redundancies, i.e., noise in the input, which is irrelevant for output prediction~\citep{TishbyIB}.

In practice, accurate MI estimation is known to be notoriously challenging, particularly for random variables of high dimension. Standard non-parametric estimators with binning, kernel density estimation (KDE), and $k$-nearest neighbor ($k$-NN)~\citep{moon1995estimation, kraskov2004estimating}, are limited by the curse of dimensionality. These estimators rely on local density estimation, which becomes unreliable in high dimensions since the data points become exponentially sparse \citep{goldfeld2020convergence, mcallester2020formal}. To alleviate this challenge, the ML community has turned to neural MI estimators \citep{duong2023diffeomorphic, cheng2020club, guo2022tight, letizia2024mutual, song2019understanding, butakov2024mutual, belghazi2018mutual, chen_neural_2025, franzese2024minde}. Despite improving estimation error in high dimensions compared to standard non-parametric models, neural MI estimators are resource-intensive, requiring substantial training epochs and hyperparameter tuning to achieve stable results. Furthermore, they are fundamentally ``black-box'' models, offering {\it little to no insight into the structural dependencies inherent within data}. 

Recently, \citet{chen_neural_2025} explored a middle ground between the extreme flexibility of neural estimators and the intuitiveness of the Gaussian copula based on \textit{vector copula} theory. Their method learns ``vector ranks'' using a neural network and then uses vector copulas to learn their joint distribution. While this method offers a robust trade-off between model capacity and data efficiency, it remains susceptible to the challenges related to neural network training outlined above. In contrast, \textit{vine copulas} \cite{joe_families_1996, bedford_probability_2001, bedford_vines-new_2002, aas_pair-copula_2009} provide an alternative framework for learning high-dimensional distributions that {\it does not rely on global function approximation but rather on dimensional decomposition}. Vine copulas break down a $d$-dimensional density into a hierarchy of $d(d-1)/2$ bivariate components, allowing for the construction of highly flexible models. \citet{nagler_evading_2016} have proposed a non-parametric implementation of the vine copula and shown that, under specific model assumptions\footnote{The assumptions are discussed in \cref{sec:guarantees}.}, the {\it density estimator converges at a rate that is independent of the dimensions, circumventing the curse of dimensionality}. This non-parametric implementation does not suffer from the costly training cycle and hyperparameter tuning of neural MI estimators.

\paragraph{Summary of contributions.}
\begin{itemize}
    \item We propose an efficient and interpretable MI estimator ``VINE'' built on non-parametric vine copulas with a closed-form expression for the density ratio under a simplified D-vine copula model.

    \item Our extensive experiments on popular datasets with known ground-truth MI values and challenging (input, output) dependence structures show that \VINE is computationally efficient and improves estimation error compared to SotA neural MI estimator baselines.
\end{itemize}
\section{Preliminaries}\label{sec:preliminaries}
\subsection{Vine copulas}
A copula $C$ with associated copula density $c$ is a $d$-dimensional distribution function with Uniform$(0,1)$ margins. By Sklar's theorem \citep{sklar1959}, any joint PDF can be characterized in terms of its marginal densities and a dependence structure, i.e., a copula:\footnote{Our notations are standard and summarized in~\cref{app:vines}.}
\begin{theorem}[Sklar's theorem]\label{thm:sklar}
    Suppose $Z_1, \dots, Z_d$ are absolutely continuous with joint PDF $f$, marginal PDFs $f_1, \dots,f_d$ and CDFs $F_1, \dots, F_d$. Then, there exists a copula $C$ with corresponding copula density $c$ such that $f(z_1, \dots, z_d) = \prod_{j=1}^d f_j(z_j) \times c(F_1(z_1), \dots, F_d(z_d))$.
\end{theorem}
Sklar's theorem implies that the marginal distributions and the joint dependence structure can be completely disentangled. Vine copulas \citep{joe_families_1996, bedford_probability_2001, bedford_vines-new_2002, aas_pair-copula_2009} are constructed by {\it decomposing the copula density into a product of $d(d-1)/2$ pair copulas}. Specifically, the model can be characterized by a sequence of $d-1$ trees, each consisting of a node set $N_t$ and an edge set $E_t$. A specific edge $e \in E_t$ represents a pair copula between two variables conditional on $t-1$ other variables. Let us denote an edge $e = (j_e, k_e; D_e)$ where $j_e, k_e$ denote the conditioned variables and $D_e$ the set of conditioning variables of edge $e$. The product of all pair copula densities constitutes the full copula density with the joint PDF of $Z_1, \dots, Z_d$ as follows: 
\begin{align}\label{eq:vinecop}\tag{VinePDF}
    f(z_1, \dots, z_d) &= \prod_{l=1}^d f_l(z_l) \times \prod_{t=1}^{d-1} \prod_{e \in E_t} c_{j_e, k_e; D_e}(F_{j_e | D_e}(z_{j_e} | \boldsymbol{z}_{D_e}), F_{k_e | D_e}(z_{k_e} | \boldsymbol{z}_{D_e})),
\end{align}
where $F_{j|D}$ denotes a conditional CDF of variable $j$ given variables with indices in the set $D$. There are several possible structures for the vine tree sequence. An example of a tree sequence in the class of D-vines is shown in \cref{fig:dvine}, where each node is connected to at most two other nodes. 

\paragraph{Vine copula's simplifying assumption.} 
We adopt the standard simplifying assumption \citep{hobaek_haff_simplified_2010, stober_simplified_2013, nagler_simplified_2025} for computational tractability. Briefly explained, under the simplifying assumption one assumes that the dependence between $Z_{j_e}$ and $Z_{k_e}$ conditional of $\boldsymbol{Z}_{D_e} = \boldsymbol{z}_{D_e}$ is invariant to the conditioning value $\boldsymbol{z}_{D_e}$, i.e., $c_{j_e, k_e; D_e}$ depends on $\boldsymbol{z}_{D_e}$ only through its two arguments, for all edges  $e = (j_e,k_e;D_e)$ that are explicit in the vine tree sequence\footnote{Importantly, this means that we do not impose the same restriction to conditional dependencies that are not directly modeled by the vine \citep{Czado2022}.}. While there are distributions where the simplifying assumption holds, such as the multivariate Gaussian distribution \citep{stober_simplified_2013}, it will generally be an approximation. Several works illustrate that simplified vines may nevertheless serve as good approximations \citep{hobaek_haff_simplified_2010, killichesSimplifying2017,spanhelSimplified2019, stober_simplified_2013,nagler_evading_2016}. For further evidence of the practical usefulness of simplified vine models, see, e.g., \citet{HukQuasiBayesVines2024,Xiao2026CopulaSVI, TagasovskaVineCopulaAutoencoders2019}. More details about the simplifying assumption are provided in \cref{app:vines}.

\subsection{Mutual information}

Consider two random vectors $\boldsymbol{X} \in \mathbb{R}^{d_X}$ and $\boldsymbol{Y} \in \mathbb{R}^{d_Y}$ with joint PDF $f$ and marginal PDFs $f_X$ and $f_Y$. The MI $I(\boldsymbol{X}; \boldsymbol{Y})$ is the Kullback-Leibler (KL) divergence between the joint distribution and the product of the marginals:
\begin{equation}\tag{MI}
I(\boldsymbol{X}; \boldsymbol{Y}) = D_{KL} \left( f \parallel f_Xf_Y \right) = \mathbb{E}_{f} \Big[ \log \frac{f(\boldsymbol{X}, \boldsymbol{Y})}{f_X(\boldsymbol{X})f_Y(\boldsymbol{Y})} \Big].
\end{equation}

Since $I(\boldsymbol{X};\boldsymbol{Y})$ depends on a typically unknown joint density function, several methods have been developed to estimate MI using finite data samples. Traditional non-parametric methods including binning, kernel density estimation (KDE) \citep{moon1995estimation}, and $k$-NN estimators \citep{kraskov2004estimating} to estimate $f$ directly, are fundamentally {\it limited by the curse of dimensionality}. To address this, the research community has developed neural MI estimators, which can be broadly categorized into deep generative models that learn explicit density parameterizations \citep{cheng2020club, franzese2024minde, mcallester2020formal, song2019understanding}, and density ratio estimation methods that utilize neural discriminators to distinguish joint samples $(\boldsymbol{X},\boldsymbol{Y} \sim f(\boldsymbol{X}, \boldsymbol{Y}))$ from marginal products $(\boldsymbol{X},\boldsymbol{Y} \sim f_X(\boldsymbol{X})f_Y(\boldsymbol{Y}) )$ \citep{belghazi2018mutual, oord2018representation, srivastava2023estimating, letizia2024mutual}. An alternative strategy involves isolating the dependency structure from the marginal distribution using copula theory. While the non-parametric method of \citet{safaai2018information} was restricted to low-dimensional settings, the recent Vector Copula Estimator (VCE) framework \citep{chen_neural_2025} extended this concept to high-dimensional vectors by adopting a vector copula structure. A more detailed explanation of MI estimators, including MINE \citep{belghazi2018mutual}, InfoNCE \citep{oord2018representation}, SMILE \citep{song2019understanding}, MIENF \citep{butakov2024mutual}, and VCE \cite{chen_neural_2025}, is given in \cref{app:mi_est}.

\section{\VINE mutual information estimator}
\label{sec:methods}
We introduce {\it \VINE, a computationally efficient simplified D-vine copula-based MI estimator}. The MI of $\boldsymbol{X}$ and $\boldsymbol{Y}$ is a function of their joint PDF $f$, and by assuming that $f$ is a simplified D-vine, the {\it density ratio $f/(f_X f_Y)$ simplifies to a product of pair copulas linking together elements of $\boldsymbol{X}$ and elements of $\boldsymbol{Y}$}. Thus, the logarithm of the density ratio becomes a sum over logarithms of pair copulas. R-vine tree structures can in principle be selected from data via Di{\ss}mann's algorithm \citep{dismann_statistical_2010} and similar methods in the non-parametric case \citep{nagler_evading_2016}. As there are $\tfrac{d!}{2} \cdot 2^{(d-2)(d-3)/2}$ possible R-vine structures for $d$ variables \citep{morales-napolesCountingVines2011}, restricting to D-vine structures may appear to sacrifice flexibility. However, we show that it leads to an {\it intuitive expression for the MI with improved estimation error in high dimensions since we do not estimate $f_X$ and $f_Y$ separately from the joint density $f$.} 

\begin{proposition}[MI for a simplified D-vine]
\label{prop:mi-for-dvine}
    Let $\boldsymbol{X} \in \mathbb{R}^{d_X}$, $\boldsymbol{Y} \in \mathbb{R}^{d_Y}$, and ${d = d_X + d_Y}$, and define $\boldsymbol{Z} = \begin{pmatrix} \boldsymbol{X}, \boldsymbol{Y} \end{pmatrix} \in \mathbb{R}^{d}$, such that 
\begin{align*}
    Z_l = 
    \begin{cases}
    X_l, & \text{if } 1 \leq l \leq d_X \\
    Y_{l - d_X}, & \text{if } d_X < l \leq d
    \end{cases}.
\end{align*} 
Suppose the joint PDF $f$ is a simplified D-vine. Then the MI reduces to
\begin{align}\tag{SumLogMI}
    I \! \left( \boldsymbol{X}; \boldsymbol{Y} \right) 
    &= \E_{f} \! \Big[ \sum_{e \in E_{XY}} \log{c_{j_e, k_e; D_e} \! \left( F_{j_e | D_e}(Z_{j_e} | \boldsymbol{Z}_{D_e}), F_{k_e | D_e}(Z_{k_e} | \boldsymbol{Z}_{D_e}) \right)} \Big],
    \label{eq:mi-dvine}
\end{align}
where $E_{XY} = \cup_{t=1}^{d-1} \{ e \in E_t \colon 1 \leq j_e \leq d_X < k_e \leq d \}$. In other words, $E_{XY}$ is the set of edges in the D-vine linking elements of $\boldsymbol{X}$ and elements of $\boldsymbol{Y}$ through a pair copula.
\end{proposition}
\begin{proof}
    See \cref{app:proofs}.
\end{proof}

\begin{figure}[htpb]
    \centering
    \begin{tikzpicture}[
    font=\Large,
    every node/.style={inner sep=1.5pt, outer sep=0pt}
]

\definecolor{fxcol}{RGB}{231,125,78}
\definecolor{fycol}{RGB}{74,151,210}
\definecolor{fxycol}{RGB}{58,183,160}
\definecolor{blobA}{RGB}{236,183,162}
\definecolor{blobB}{RGB}{168,203,232}
\definecolor{blobC}{RGB}{157,221,208}

\node at (-5.6, 2.55) {$f$};
\node at (-4.3, 2.55) {$=$};


\fill[blobA, rounded corners=20pt, fill opacity=0.6]
(-4.00, 2.95) -- (3.20, 2.95) -- (-0.40, -0.20) -- cycle;

\fill[blobB, rounded corners=20pt, fill opacity=0.6]
(3.80, 2.95) -- (11.0, 2.95) -- (7.40, -0.20) -- cycle;

\fill[blobC, rounded corners=20pt, fill opacity=0.6]
(3.50, 2.15) -- (-0.25,  -0.55) -- (3.50, -3.25) -- (7.05,  -0.55) -- cycle;

\node at (-3.00, 2.55) {$f_1$};
\node at (-1.70, 2.55) {$\times$};
\node at (-0.40, 2.55) {$f_2$};
\node at (0.90, 2.55) {$\times$};
\node at (2.20, 2.55) {$f_3$};
\node at (3.50, 2.55) {$\times$};
\node at (4.80, 2.55) {$f_4$};
\node at (6.10, 2.55) {$\times$};
\node at (7.40, 2.55) {$f_5$};
\node at (8.70, 2.55) {$\times$};
\node at (10.00,2.55) {$f_6$};

\node at (-2.95, 1.55) {$\times$};
\node at (-1.70, 1.55) {$c_{1,2}$};
\node at (-0.45, 1.55) {$\times$};
\node at (0.90, 1.55) {$c_{2,3}$};
\node at (2.15, 1.55) {$\times$};
\node at (3.50, 1.55) {$c_{3,4}$};
\node at (4.75, 1.55) {$\times$};
\node at (6.10, 1.55) {$c_{4,5}$};
\node at (7.35, 1.55) {$\times$};
\node at (8.70, 1.55) {$c_{5,6}$};

\node at (-1.75, 0.55) {$\times$};
\node at (-0.45, 0.55) {$c_{1,3;2}$};
\node at (0.85,  0.55) {$\times$};
\node at (2.15,  0.55) {$c_{2,4;3}$};
\node at (3.45,  0.55) {$\times$};
\node at (4.75,  0.55) {$c_{3,5;4}$};
\node at (6.05,  0.55) {$\times$};
\node at (7.35,  0.55) {$c_{4,6;5}$};

\node at (-0.45, -0.55) {$\times$};
\node at (0.85,  -0.55) {$c_{1,4;2,3}$};
\node at (2.15,  -0.55) {$\times$};
\node at (3.45,  -0.55) {$c_{2,5;3,4}$};
\node at (4.75,  -0.55) {$\times$};
\node at (6.05,  -0.55) {$c_{3,6;4,5}$};

\node at (0.85,  -1.55) {$\times$};
\node at (2.15,  -1.55) {$c_{1,5;2,3,4}$};
\node at (3.45,  -1.55) {$\times$};
\node at (4.75,  -1.55) {$c_{2,6;3,4,5}$};

\node at (2.15,  -2.55) {$\times$};
\node at (3.45,  -2.55) {$c_{1,6;2,3,4,5}$};

\node[fxcol] at (-3.65,0.2) {$f_X$};
\draw[fxcol,line width=1pt]
(-3.15, 0.2) .. controls (-3.0,0.32) and (-2.5,0.3) .. (-2.2, 1.0);

\node[fycol] at (12.5, 3.0) {$f_Y$};
\draw[fycol,line width=1pt]
(10.9, 2.75) .. controls (11.30,3.15) and (11.6,2.95) .. (12.00,3.05);

\node[fxycol] at (6.7,-3.0) {$\dfrac{f}{f_X f_Y}$};
\draw[fxycol,line width=1pt]
(5.0, -2.55) .. controls (5.4, -2.8) and (5.6, -2.9) .. (6.0, -3.0);

\end{tikzpicture}
    \caption{Identifying the density ratio from a D-vine density. Functional arguments are omitted to simplify notation.}
    \label{fig:dvine-decomp}
\end{figure}
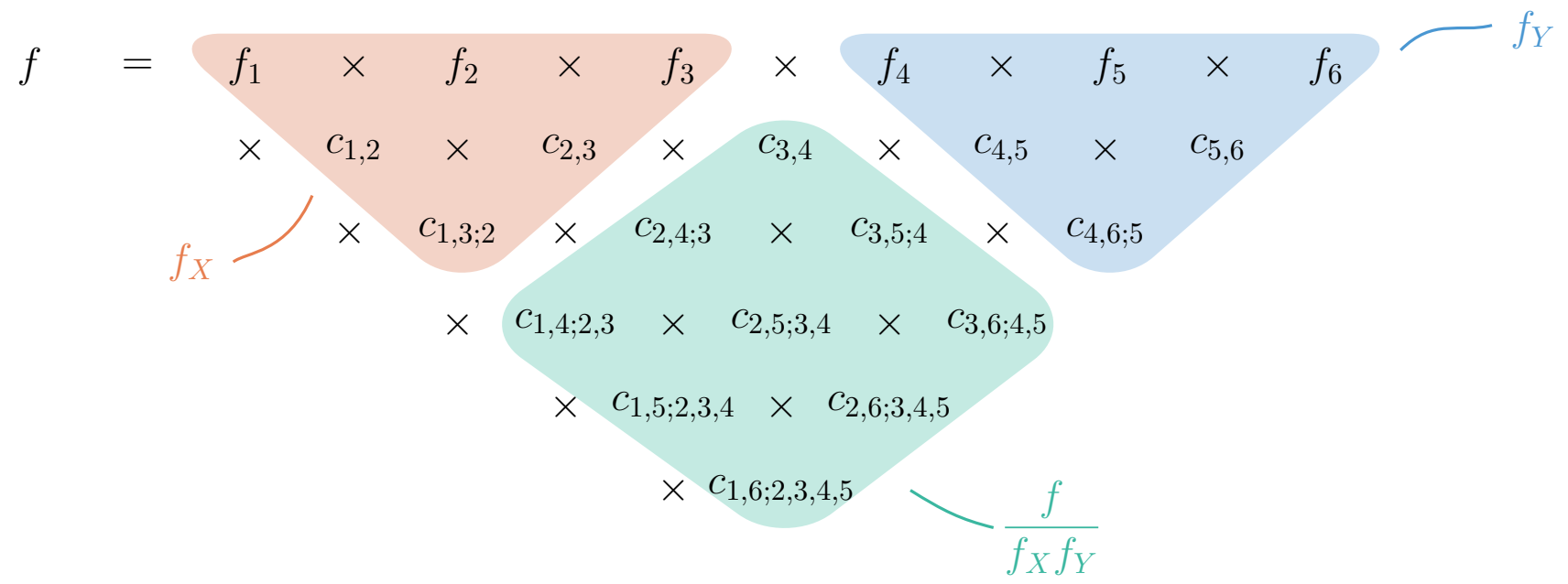

To illustrate this construction, we show a simple example. Let $d=6$ and $d_X = d_Y = 3$, such that $(Z_1, Z_2, Z_3) = (X_1, X_2, X_3)$ and $(Z_4, Z_5, Z_6) = (Y_1, Y_2, Y_3)$. The full joint PDF is visualized in \cref{fig:dvine-decomp}, where we have highlighted $f_X$ in orange and $f_Y$ in blue. From this figure, we see how the density ratio $f/(f_X f_Y)$, being the remaining terms that are highlighted in green, becomes the product of the pair copulas linking elements of $\boldsymbol{X}$ and elements of $\boldsymbol{Y}$. 

\VINE leverages on~\cref{prop:mi-for-dvine} to estimate the full joint PDF (including the margins and the copula) and extracts the relevant pair copulas to compute a Monte Carlo (MC)-type estimate. Given $n$ i.i.d.\ samples from the true joint density $f$, $\{\boldsymbol{Z}^{(i)}\}_{i=1}^n = \{(\boldsymbol{X}^{(i)}, \boldsymbol{Y}^{(i)})\}_{i=1}^n$, we estimate:
\begin{align}\tag{\VINE}
    \hat{I}(\boldsymbol{X},\boldsymbol{Y}) = \frac{1}{n} \sum_{i=1}^n \sum_{e \in E_{XY}} \log{\hat{c}_{j_e,k_e;D_e} \! \left( \hat{F}_{j_e|D_e}(Z_{j_e}^{(i)} | \boldsymbol{Z}_{D_e}^{(i)}), \hat{F}_{k_e|D_e}(Z_{k_e}^{(i)} | \boldsymbol{Z}_{D_e}^{(i)}) \right)}.
    \label{eq:mi-dvine-estimate}
\end{align}
The general procedure is summarized in~\cref{alg:mi-vinecop}. In principle, the different components of \VINE can be estimated both parametrically and non-parametrically. However, we choose a fully non-parametric approach, as proposed by \citet{nagler_evading_2016}. In~\cref{sec:guarantees}, we argue theoretically why a fully non-parametric implementation of our proposed MI estimator is ideal.

\begin{algorithm}
\caption{\VINE: a simplified D-vine copula based MI estimator.}
\label{alg:mi-vinecop}
\begin{algorithmic}[1]
\REQUIRE Data $\{ \boldsymbol{Z}^{(i)} \}_{i=1}^n$ with $\boldsymbol{Z}^{(i)} = (\boldsymbol{X}^{(i)}, \boldsymbol{Y}^{(i)})$ and $\boldsymbol{X}^{(i)} \in \mathbb{R}^{d_X}, \boldsymbol{Y}^{(i)} \in \mathbb{R}^{d_Y}$
\ENSURE MI estimate $\hat{I}(\boldsymbol{X}; \boldsymbol{Y})$
\STATE Estimate marginal CDFs $\hat{F}_1, \dots, \hat{F}_d$.
\STATE Transform data to pseudo-observations $\hat{F}_{l}(Z_{l}^{(i)})$, as described in \citet{czado_analyzing_2019}. 
\STATE Estimate a D-vine copula using the pseudo-observations to obtain pair copula estimates
\begin{align*}
    \{ \hat{c}_{j_e, k_e; D_e} : e \in E_t, t = 1, \dots, d-1 \}.
\end{align*}
\STATE Compute $\hat{I}(\boldsymbol{X}; \boldsymbol{Y})$ using \cref{eq:mi-dvine-estimate}.
\end{algorithmic}
\end{algorithm}

The flexibility of \VINE can be controlled using tuning parameters. A common strategy for pruning the model is to set selected pair copulas to independence copulas. We employ \textit{thresholding} \citep{naglerModelselectionsparse2019}, which means setting pair copulas that exhibit weak empirical dependencies to independence copulas. Specifically, an edge is pruned if its absolute empirical Kendall's $\tau$ falls below a prespecified threshold. This approach allows the estimator to effectively resolve sparse dependence structures while reducing the risk of overfitting. Additionally, to improve the computational speed of the estimator, we parallelize the estimation of individual pair copulas within each tree level. \VINE is implemented using the Python package \texttt{pyvinecopulib} \citep{naglerpyvinecopulib2025}.
\section{Theoretical motivation}
\label{sec:guarantees}
Suppose we have i.i.d. data $\{\textbf{Z}^{(i)}\}_{i=1}^n = \{(\textbf{X}^{(i)}, \textbf{Y} ^{(i)})\}_{i=1}^n$ sampled from a simplified D-vine density $f$ as defined in \cref{sec:preliminaries}. As shown in \cref{sec:methods}, the true MI simplifies to the expectation of the sum of logarithms of pair copulas linking together the elements of $\textbf{X}$ and $\textbf{Y}$. Even in an ideal setting where all components of $f$ are known, we might still \textit{have to approximate the expectation by MC}, especially when $d$ is large. An ideal MI estimator would therefore be
\begin{equation}
    \tag{IdealMIEst}
    I^*(\textbf{X}, \textbf{Y}) = \frac{1}{n} \sum_{i=1}^n \sum_{e\in E_{XY}} \log c_{j_e, k_e; D_e} \left(F_{j_e|D_e} (Z^{(i)}_{j_e} | \textbf{Z}^{(i)}_{D_e}), F_{k_e | D_e}(Z^{(i)}_{k_e} |\textbf{Z}^{(i)}_{D_e})\right).
    \label{eq:ideal_mi}
\end{equation}
Under regularity conditions\footnote{We require that $\text{Var}\left( \sum_{e\in E_{XY}} \log c_{j_e, k_e; D_e} \left(F_{j_e|D_e} (Z^{(i)}_{j_e} | \textbf{Z}^{(i)}_{D_e}, F_{k_e | D_e}(Z^{(i)}_{k_e} |\textbf{Z}^{(i)}_{D_e})\right) \right) < \infty$.}, \cref{eq:ideal_mi} converges in probability to the ground-truth $I(\textbf{X}, \textbf{Y})$ by the Weak Law of Large Numbers. Realistically, all components of $f$ are unknown and must be estimated from the data. If $\hat f$ is the D-vine density estimator, then plugging in its components into \cref{eq:ideal_mi} yields the proposed estimator in \VINE. Therefore, if $\hat f$ is a consistent approximation of $f$, \VINE becomes a good approximation of \cref{eq:ideal_mi}, which is a good approximation of the true MI with a large enough amount of data. 

There are several motivations behind estimating the components of the simplified vine model non-parametrically. A fully non-parametric implementation of \VINE avoids model misspecification and offers higher flexibility than a parametric implementation. Moreover, \citet{nagler_evading_2016} have shown that under certain assumptions, non-parametric vine density estimators achieve {\it convergence rates that are independent of the ambient dimension}. Specifically, if the pair copulas are $K$ times differentiable, and $\hat{f}$ satisfies assumptions A1 to A4 in \citet{nagler_evading_2016}, then for a given $\boldsymbol{z}$, $|\hat{f}(\boldsymbol{z}) - f(\boldsymbol{z})|$ is bounded in probability at rate $n^{-r}$, where $r=K/(2K+2)$. In other words, this non-parametric density estimator overcomes the ``curse of dimensionality'' faced by essentially all consistent $d$-dimensional non-parametric density estimators, where the error rate increases exponentially with the ambient dimension \citep{stone1980optimal}. Thus, under the simplifying and D-vine assumptions, we expect \VINE to outperform MI estimators based on full-dimensional non-parametric density estimation (e.g. binning, KSG \citep{kraskov2004estimating}, KDE \citep{moon1995estimation}).

Even in settings where the simplifying assumption and D-vine assumption do not hold, it may still be the case that $f$ can be well approximated by a simplified D-vine density $\hat{f}$. Furthermore, although the estimation of $\hat{f}$ involves $d + d(d-1)/2$ estimation tasks, training \VINE may still be less computationally demanding than training a full neural network.

\section{Numerical experiments}
\label{sec:experiments}

In this section, we evaluate \VINE's performance on a range of benchmark datasets with known ground-truth MI. We examine how dimensionality, dependency structures, and data structures affect \VINE's performance, and consider settings where the assumptions underlying \VINE hold and settings where they are known to be violated. In all experiments, we use a fully non-parametric D-vine estimator and fix the threshold parameter $\tau=0.1$. We benchmark \VINE against SotA neural and vector copula based MI estimators including MINE, InfoNCE, MIENF, SMILE, and VCE.

\subsection{Evaluating the simplifying assumption} \label{sec:exp_simplifying}
\textbf{Experimental setup.} We evaluate the estimators under four data-generating distributions spanning settings in which the simplifying assumption holds and settings in which it is violated. In the former, we generate $(\bf{X}, \bf{Y}) \sim N_d(\bf{0}, \bf{\Sigma})$, with two different covariance structures: pairwise dependence only between $X_i$ and $Y_i$, and spatial dependence, defined by $\text{corr}(X_i, Y_j) = \exp ( - \text{dist}(i,j) / \rho )$, where $\text{dist}(i,j)$ denotes the distance between the coordinates of the features on a 2D grid. For the latter, we use a Mixture of Gaussians (MoG) distribution configured with the same covariance structures described above. We expect the mixture with sparse, pairwise-only dependence to exhibit a milder violation, and the spatial-dependence mixture to produce a stronger violation of the simplifying assumption.

\begin{figure}[tb]
    \centering
    \begin{subfigure}[b]{0.32\textwidth}
        \centering
        \includegraphics[width=\textwidth]{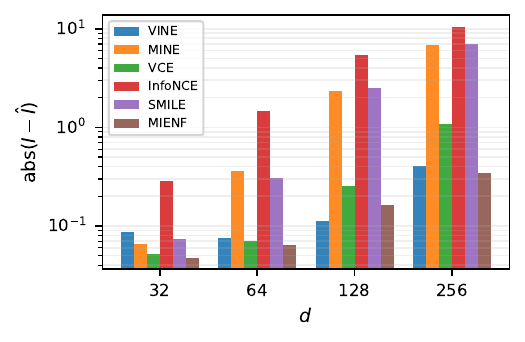}
        \caption{Gaussian Pairs}
        \label{fig:sa_gauss_pairs}
    \end{subfigure}%
    \hspace{0.05\textwidth}
    \begin{subfigure}[b]{0.32\textwidth}
        \centering
        \includegraphics[width=\textwidth]{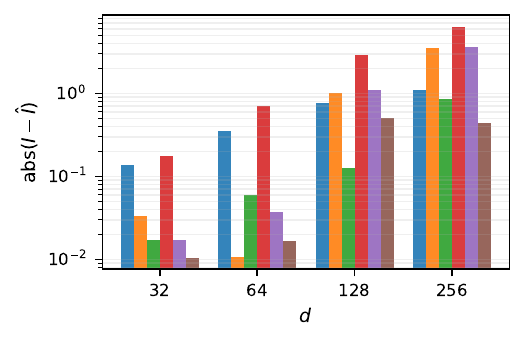}
        \caption{Gaussian Spatial}
        \label{fig:sa_gauss_spatial}
    \end{subfigure}

    \vspace{2mm} 

    \begin{subfigure}[b]{0.32\textwidth}
        \centering
        \includegraphics[width=\textwidth]{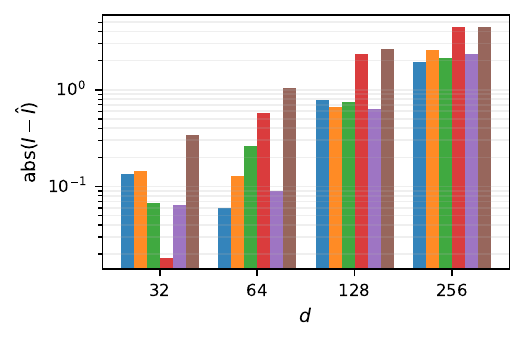}
        \caption{MoG Pairs}
        \label{fig:sa_mog_pairs}
    \end{subfigure}%
    \hspace{0.05\textwidth}
    \begin{subfigure}[b]{0.32\textwidth}
        \centering
        \includegraphics[width=\textwidth]{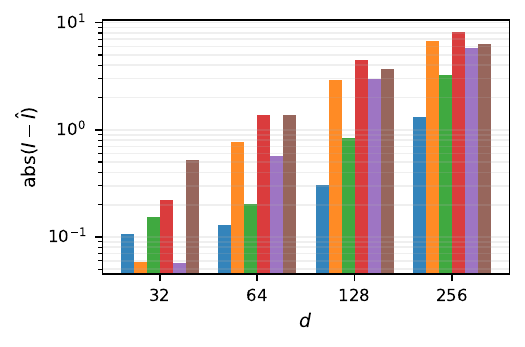}
        \caption{MoG Spatial}
        \label{fig:sa_mog_spatial}
    \end{subfigure}
    \caption{Performance of \VINE and neural estimators across data distributions where the simplifying assumption holds (Gaussian) versus where it is violated (Mixture of Gaussians).}
    \label{fig:simplifying_assumption}
\end{figure}

\textbf{Results.} The results are summarized in \cref{fig:simplifying_assumption}. In the Gaussian task, where the simplifying assumption holds, \VINE demonstrates high accuracy and remains robust to high dimensionality and ground-truth MI compared to most neural estimators. We especially observe that MINE, InfoNCE, and SMILE suffer substantial error degradation for $d=128$ and $d=256$, while \VINE performs similarly to or better than the best neural estimators. Crucially, in the MoG dataset where the simplifying assumption is violated, \VINE remains remarkably accurate and competitive against state-of-the-art neural estimators. This indicates that while the simplifying assumption is necessary for theoretical bounds, its violation in practice does not critically break the D-vine's ability to approximate the global dependency structure. Furthermore, \VINE achieves this with significant computational efficiency, computing estimates in a fraction of the time required for neural training (see \cref{fig:scaling_limitations}). \VINE also remains accurate in high dimensions (up to $d=256$). For a more thorough investigation of \VINE's limitations in terms of dimensionality and computational time and accuracy, see \cref{app:limitations}.

\subsection{Non-Gaussian datasets}
\textbf{Experimental setup.} 
Following \citet{mi_benchmarking_NEURIPS2023}, we extend MI evaluation beyond the simple multivariate Gaussian case. We consider cases where both model assumptions underlying \VINE hold true and are violated:
\begin{itemize}
    \item \textbf{Margin-wise maps of elliptical distributions}: Elliptical distributions, such as the multivariate Gaussian and the multivariate Student's t, have copulas satisfying the simplified D-vine assumption \cite[Theorem~4.1]{stober_simplified_2013}. According to the invariance property of copulas \cite[p.~18]{czado_analyzing_2019}, applying strictly increasing transformations to the margins of these distributions preserve the underlying copula, and thus keep model assumptions intact. We consider multivariate Gaussian data with marginal transformations $g(x)=\sign{x} \cdot x^2, g(x)=x^3, g(x)=e^x$, and multivariate Student's t data.
    \item \textbf{Cross-dimensional maps of Gaussian distributions}: \citet{mi_benchmarking_NEURIPS2023} consider cross-dimensional maps of Gaussian variables, such as the Spiral diffeomorphism and the Swiss Roll data. Although these transformations preserve the true MI, they do not preserve the underlying copula, so the simplified D-vine assumption may no longer hold.
    \item \textbf{Mixture of Gaussian distributions}: Mixture of Gaussian (MoG) distributions can be expressed as $(\boldsymbol{X}, \boldsymbol{Y}) \vert C=k \sim \mathrm{N}_d(\boldsymbol{\mu}_k, \boldsymbol{\Sigma}_k)$, for $k=1,\dots,K$, and where $C$ is a latent variable. MoG data do not satisfy the simplified D-vine assumption.
\end{itemize}
Each experiment is run using $10,000$ data points. See \cref{app:transformed_gaussian} for further details of the data generating distributions and comments on the simplifying assumption. 

\begin{figure*}[tb]
    \centering
    \footnotesize
    \setlength{\tabcolsep}{2pt}
    
    \begin{tabular*}{\textwidth}{@{\extracolsep{\fill}}ccc}
        
        \includegraphics[width=0.32\textwidth]{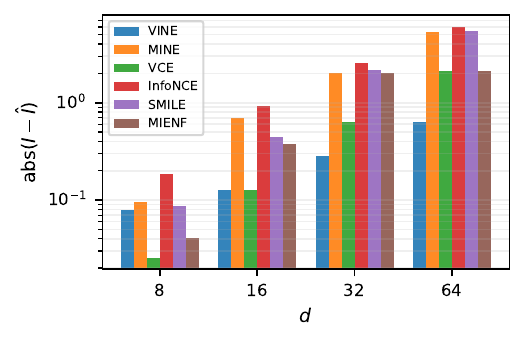} &
        \includegraphics[width=0.32\textwidth]{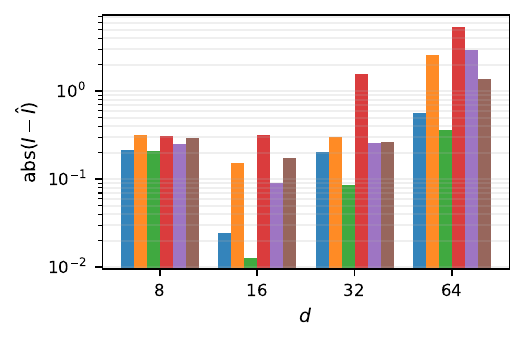} &
        \includegraphics[width=0.32\textwidth]{images/t_r1_abs_error_mog.pdf} \\
        (a) $\text{sign}(X)\cdot X^2, e^Y$ & (b) Spiral & (c) MoG \\ [2mm]

        \includegraphics[width=0.32\textwidth]{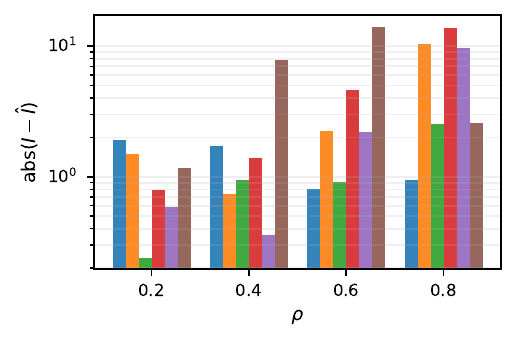} &
        \includegraphics[width=0.32\textwidth]{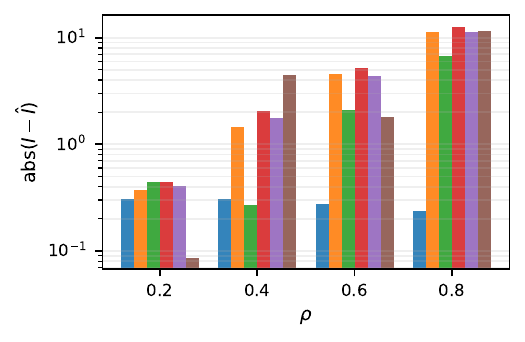} &
        \includegraphics[width=0.32\textwidth]{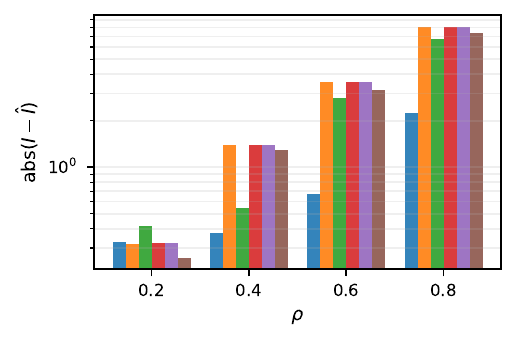} \\
        (d) Student's-$t$, $\nu=2$ & (e) $X^3, Y^3$ & (f) Swiss Roll \\
        
    \end{tabular*}

    \caption{\textbf{MI Estimation on transformed and non-Gaussian data.} Row 1 shows the absolute estimation error $|I - \hat{I}|$ as dimension $d$ increases (fixed $\rho=0.5$). Row 2 displays error across varying correlation strengths $\rho$ for complex non-linear dependencies (fixed $d=32$).}
    \label{fig:gaussian_transformed}
\end{figure*}

\textbf{Results.} Across all benchmarks, {\it \VINE consistently demonstrates high accuracy and stability compared to the neural estimators, especially as dimensionality and MI increase}, as seen in \cref{fig:gaussian_transformed}. For marginal transformations such as cube and sign-square mappings, the performance gap is especially pronounced: the vine's rank-based mapping inherently exploits copulas' invariance property and recovers the underlying dependency structure. The neural critics, e.g., MINE and InfoNCE struggle to recover such structure. We note that the simplifying assumption holds in these cases. In more geometrically complex tasks, such as the Spiral and Swiss Roll data, where the simplifying assumption may not theoretically hold, the results show that \VINE still captures the vast majority of the information signal. This suggests that \VINE can be effective even when the model assumptions are violated. {\it The neural estimators exhibit poor performance in high dimensions and high dependency settings}, as demonstrated in \cref{sec:exp_simplifying}. In contrast, \VINE remains significantly more robust across these regimes. We observe that \VINE performs less optimally on the heavy-tailed Student's-$t$ data, as it overestimates ground-truth MI, likely by fitting noise within the distribution's tails. However, this sensitivity to tail noise could be mitigated by adjusting the threshold parameter, $\tau$, upward to force sparser dependencies. While the VCE remains exceptionally accurate on the MoG dataset, likely aided by its architectural assumption of a Gaussian mixture structure within the vector copula, \VINE performs well on datasets requiring higher structural flexibility, such as Swiss Roll and the cubed Gaussian.

\subsection{Image dataset with known MI} \label{sec:exp_image}
\begin{figure*}[t]
    \centering
    \begin{subfigure}{0.23\textwidth}
        \centering
        \small $X$ \\
        \includegraphics[width=\linewidth]{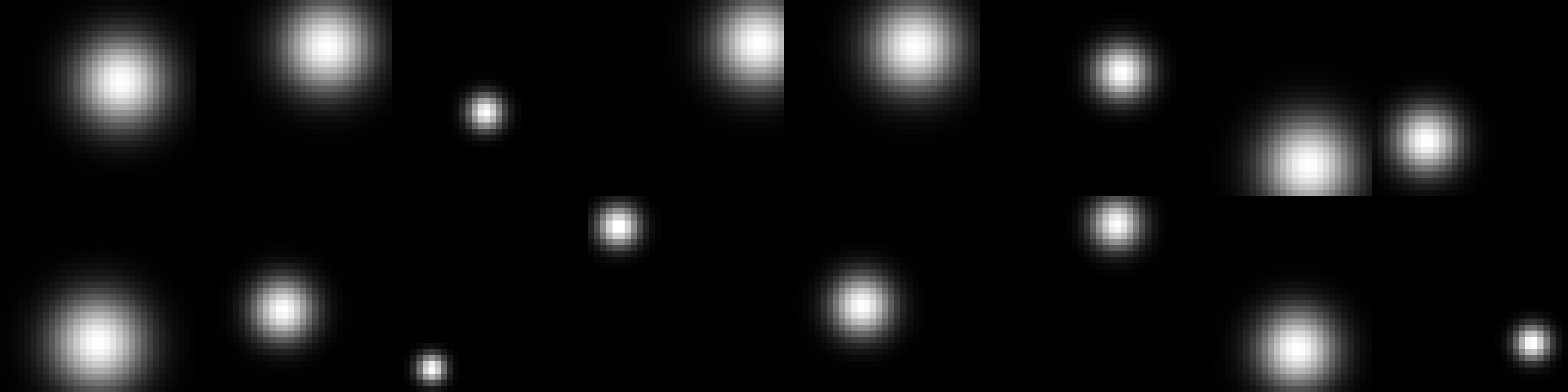}
        \vspace{-5pt} 
        \caption{} 
    \end{subfigure}
    \hfill
    \begin{subfigure}{0.23\textwidth}
        \centering
        \small $Y$ \\
        \includegraphics[width=\linewidth]{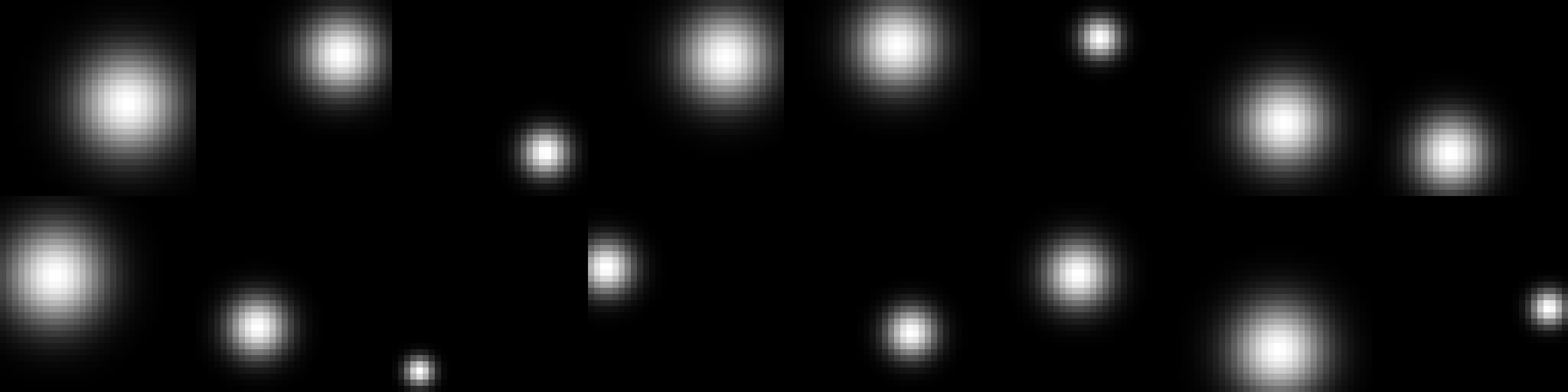}
        \vspace{-5pt}
        \caption{}
    \end{subfigure}
    \hfill
    \begin{subfigure}{0.23\textwidth}
        \centering
        \small $X$ \\
        \includegraphics[width=\linewidth]{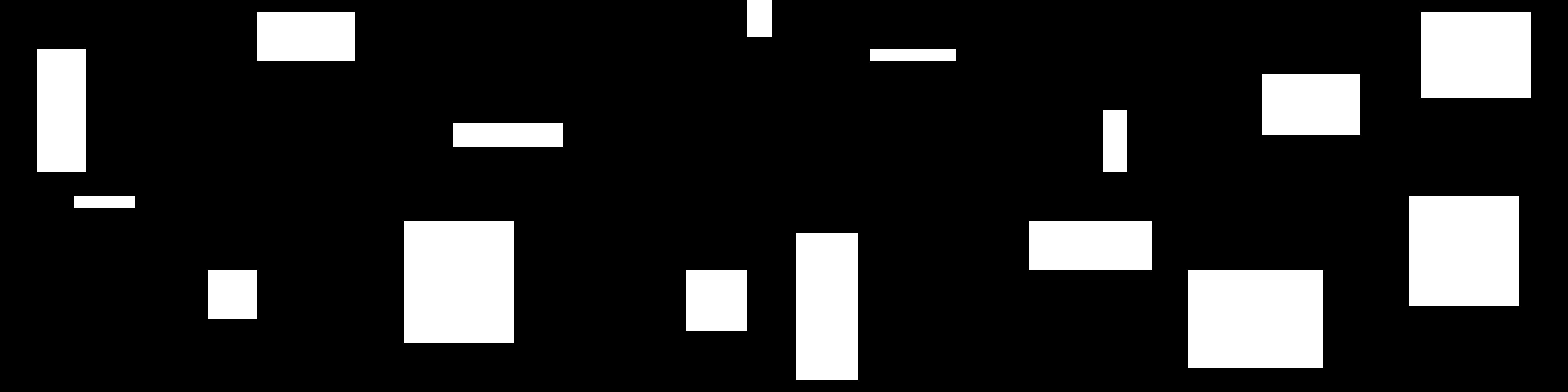}
        \vspace{-5pt}
        \caption{}
    \end{subfigure}
    \hfill
    \begin{subfigure}{0.23\textwidth}
        \centering
        \small $Y$ \\
        \includegraphics[width=\linewidth]{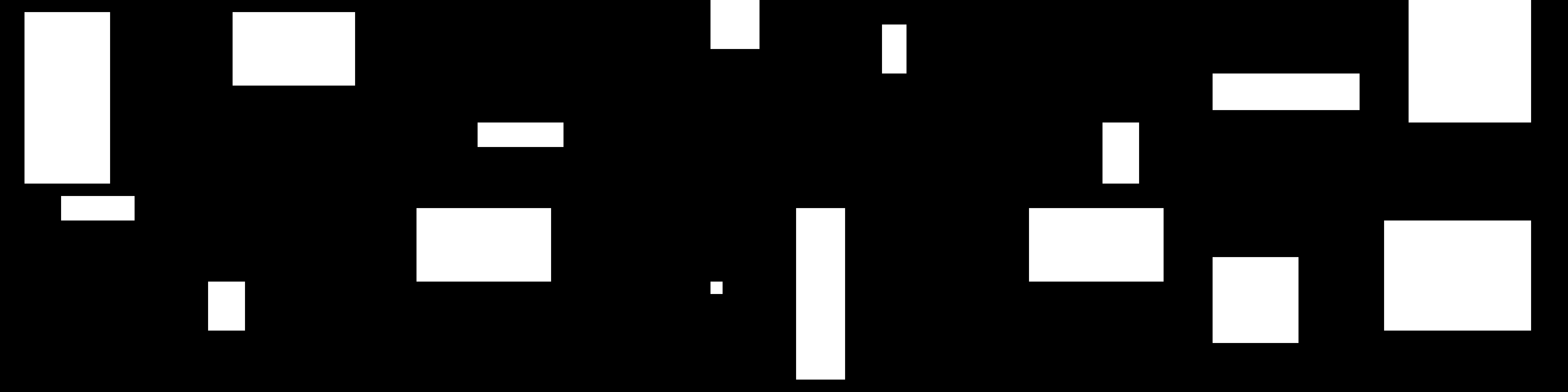}
        \vspace{-5pt}
        \caption{}
    \end{subfigure}

    \caption{Visual representation of the generated image datasets: (a-b) Gaussian plates for $X$ and $Y$, and (c-d) Rectangles for $X$ and $Y$.}
    \label{fig:all_datasets_row}
\end{figure*}

\textbf{Experimental setup.} To evaluate \VINE's ability to model high-dimensional data, we replicate the image-based benchmarks proposed by \citet{butakov2023information}. We consider two distinct synthetic image datasets: Gaussian images and rectangle images with known ground-truth MI, as visualized in \cref{fig:all_datasets_row}. MI estimation is known to be difficult in high-dimensional data: for any distribution-free MI estimator, the required sample complexity is proven to depend exponentially on the dimension of the data \citep{goldfeld2020convergence, mcallester2020formal}. However, according to the \textit{Manifold Hypothesis}, high-dimensional data, such as large images, often lie on a significantly lower-dimensional manifold. In our experiments, we consider $\boldsymbol{X} \in \R^{16 \times 16}$ and $\boldsymbol{Y} \in \R^{16 \times 16}$ images where the ground-truth $I(\boldsymbol{X};\boldsymbol{Y})$ is known. Following \citet{butakov2023information}, we preprocess the raw image data using Principal Component Analysis (PCA) $e: \R ^{16 \times 16} \rightarrow \R ^d$. We show in \cref{app:image} that PCA successfully reduces the data dimensionality while preserving information. We set $d$ to $12$ and $40$ for the Gaussian and rectangle images, respectively. We estimate MI on the latent variables with \VINE, and compare against neural estimators MINE, MINDE, and VCE. We used a training sample size of 10,000.

\begin{table}[t]
\centering
\caption{MI Estimation Performance: Gaussian Images ($d=12$) vs. Rectangles ($d=40$). Results reported as Mean ($\pm$ Std). Time indicates the average duration per estimation.}
\label{tab:image_results}
\footnotesize
\begin{tabular}{@{}llcccc@{}}
\toprule
\textbf{Dataset} & \textbf{Estimator} & $I(\boldsymbol{X};\boldsymbol{Y}) = 1$ & $I(\boldsymbol{X};\boldsymbol{Y}) = 3$ & $I(\boldsymbol{X};\boldsymbol{Y}) = 7$ & \textbf{Avg. Time} \\ \midrule

\multirow{4}{*}{\rotatebox[origin=c]{90}{\textbf{Gaussian}}} 
& MINE    & \textbf{0.992} ($\pm$0.017) & 2.848 ($\pm$0.028) & 5.802 ($\pm$0.039) & 16.49s \\
& InfoNCE & 0.950 ($\pm$0.031) & 2.604 ($\pm$0.014) & 3.823 ($\pm$0.010) & 465.8s \\
& VCE     & 0.027 ($\pm$0.051) & \textbf{3.140} ($\pm$0.096) & \textbf{7.021} ($\pm$0.021) & 124.60s \\
& \textbf{\VINE (Ours)} & 1.089 ($\pm$0.055) & 3.540 ($\pm$0.217) & 6.325 ($\pm$0.412) & \textbf{8.82s} \\ 

\addlinespace[0.3em] 
\midrule 

\multirow{4}{*}{\rotatebox[origin=c]{90}{\textbf{Rectangles}}} 
& MINE    & 0.874 ($\pm$0.043) & 2.601 ($\pm$0.066) & 5.727 ($\pm$0.163) & 86.16s \\
& InfoNCE & 0.860 ($\pm$0.040) & 2.427 ($\pm$0.057) & 3.576 ($\pm$0.122) & 633.21s \\
& VCE     & 0.100 ($\pm$0.171) & 1.282 ($\pm$0.107) & 3.747 ($\pm$0.169) & 418.35s \\
& \textbf{\VINE (Ours)} & \textbf{1.100 ($\pm$0.199)} & \textbf{3.390} ($\pm$0.254) & \textbf{7.883 ($\pm$0.340)} & \textbf{50.82s} \\ 
\addlinespace[0.3em]
\bottomrule
\end{tabular}
\end{table}

\textbf{Results and analysis.} \cref{tab:image_results} compares the performance of the MI estimators across four different levels of MI for the Gaussian and rectangle image datasets, respectively. \VINE shows high accuracy across most MI levels. Notably, in the ``high'' and ``very high'' MI regimes, standard neural estimators like MINDE and MINE exhibit significant downward bias, while \VINE remains more stable. Although the methodology proposed by \citet{butakov2023information} successfully utilized non-parametric estimators like KDE and WKL, their efficacy was strictly limited to extremely low-dimensional latent spaces ($d=4$). In our investigation, KDE and WKL at $d=12$ and $d=40$ are highly inaccurate due to the curse of dimensionality. \VINE, however, remains robust to these dimensions. This allows us to utilize robust non-parametric estimates on relatively high-dimensional latent spaces. Despite the quadratic scaling of the number of pairs, \VINE remains computationally competitive. In the Gaussian experiment, it outperforms all neural methods in training efficiency. Furthermore, in the rectangle experiment, \VINE achieves comparable speed to the fastest neural models. 

\subsection{Post-hoc inspection}

\begin{figure*}[tb]
    \centering
    \begin{subfigure}[b]{0.48\textwidth}
        \centering
        \includegraphics[width=\textwidth]{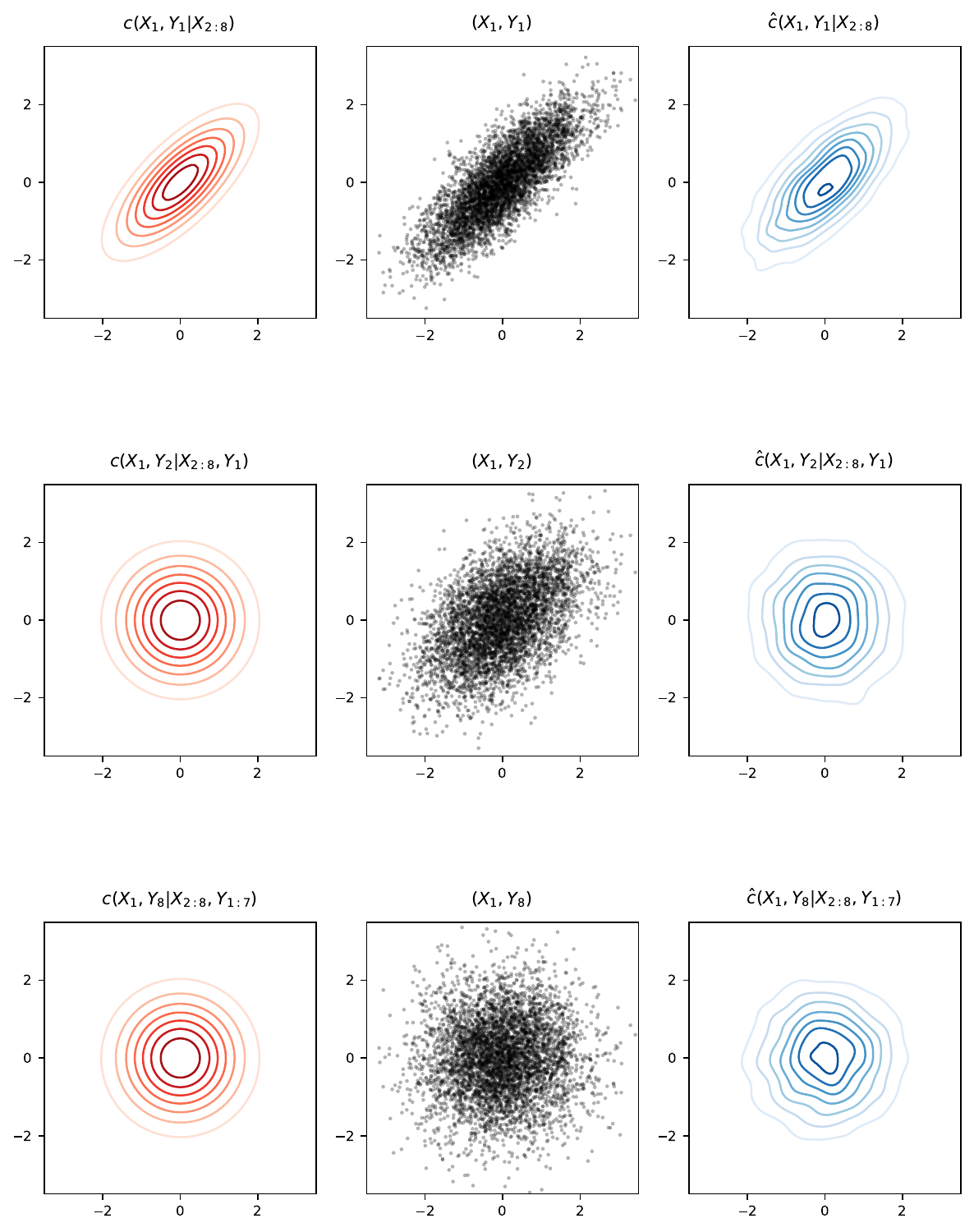}
        \caption{Gaussian Baseline}
        \label{fig:audit_gaussian}
    \end{subfigure}
    \hfill 
    \begin{subfigure}[b]{0.48\textwidth}
        \centering
        \includegraphics[width=\textwidth]{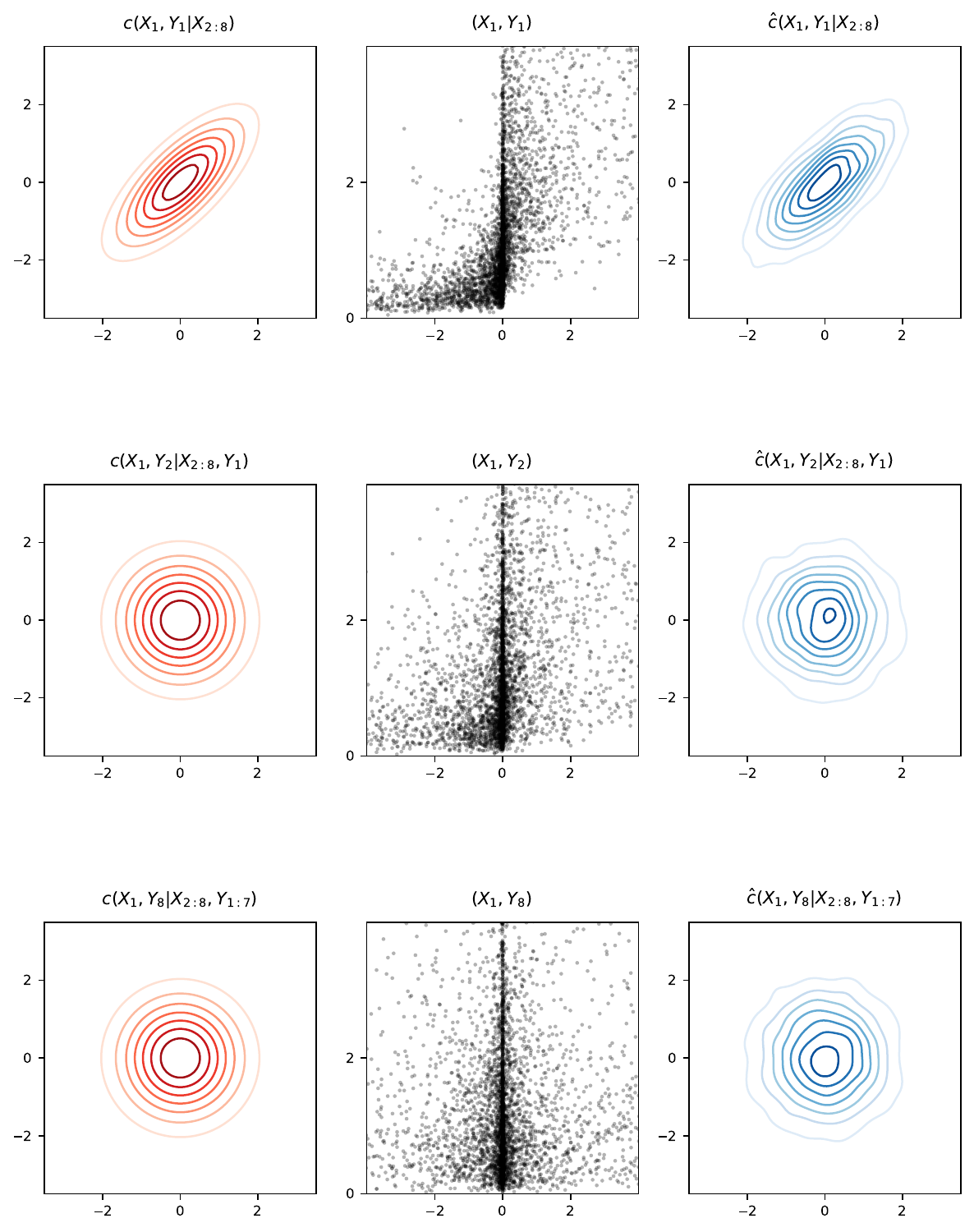}
        \caption{Transformed Data: ($X^3, \exp(Y)$)}
        \label{fig:audit_transformed}
    \end{subfigure}

    \caption{\textbf{Visualization of a selection of \VINE's learned pair copulas.} \VINE trained on Gaussian data of $d_X=d_Y=8$ (\cref{fig:audit_gaussian}) and transformed Gaussian data $X^3, \exp(Y)$ (\cref{fig:audit_transformed}). Each subfigure displays a grid comparing the ground-truth pair copula (left column), scatter plots of the raw data (center column), and the estimated pair copula (right column).}
    \label{fig:vine_audit_main}
\end{figure*}

\paragraph{Experimental setup.}

We demonstrate the structural transparency of \VINE using contour plots of the learned pair copulas. Unlike neural MI estimators, which output a single scalar from a black-box neural net, \VINE explicitly decomposes the total MI into a tractable sum of bivariate pair copula terms, directly reflecting the tree-level decomposition of the joint density (\cref{fig:dvine-decomp}). This sum allows us to identify pair copulas that contribute zero (independence copulas) and those that highly influence the final MI estimate. The highly influential pair copulas can further be visually inspected by contour plots. To demonstrate this transparency, we simulate Gaussian data ($d_X=d_Y=8$) with spatial dependence between dimensions (as explained in \cref{sec:exp_simplifying}), estimate a simplified D-vine copula, and compare the estimated bivariate copulas to the ground-truth. We also perform an experiment with the marginal transformations $X_l^3, \exp(Y_l)$. Crucially, while these transformations warp the raw scatter plots, the underlying copula structure, and thus the true MI, remains invariant. Visualizations of copulas are projected from $[0,1]^2$ to Gaussian space using the inverse CDF mapping ($\Phi^{-1}$).

\paragraph{Results.}
\cref{fig:vine_audit_main} visualizes estimated bivariate copulas next to the ground truths, along with a scatter plot of the raw data. We observe that the contour plots reveal which bivariate copulas contribute to the MI estimate, and which do not (independence copulas). For instance, the contour plot for $\hat c(X_1, Y_1 | X_{2:8})$ in \cref{fig:audit_gaussian} (top right) shows that the bivariate copula between $X_1$ and $Y_1$ given $X_2, ..., X_d$ has a strong positive correlation, indicating a significant contribution to the total MI. In contrast, the contour plots of $\hat c(X_1, Y_2 | X_{2:8}, Y_1)$ and $\hat c(X_1, Y_8 | X_{2:8}, Y_{1:7})$ (middle right and bottom right), takes the form of an independence copula with spherical contours, signaling that it contributes zero to the MI estimate. Note that although the scatter plot of $(X_1,Y_2)$ shows a positive correlation, $\hat c(X_1, Y_2 | X_{2:8}, Y_1)$ is an independence copula because $X_1$ and $Y_2$ are \textit{conditionally} independent given $X_2, ..., X_8, Y_1$. Furthermore, since the ground-truth bivariate copulas are known for simulated Gaussian data, we can visually verify that the estimated copulas closely match the ground truth. Crucially, even when non-linear transformations obscure the dependencies in the raw scatter plots (\cref{fig:audit_transformed}), the estimated bivariate copulas continue to closely match the ground truth. Finally, we note that when the simplifying assumption is violated, the learned pair-copulas represent a structured approximation of the true conditional dependencies rather than an exact recovery \cite{nagler_evading_2016, spanhelSimplified2019}.

\section{Limitations and future work}
\label{sec:limitations}
While \VINE scales well in terms of estimation error as the dimensions grow, it still requires estimating $O(d^2)$ bivariate copulas sequentially. Addressing this quadratic complexity is a remaining computational challenge. Moreover, the convergence results for the density estimator underlying \VINE depend on the well-known and common simplifying assumption. Further work should investigate how severely this assumption can be violated before performance degrades noticeably. Additional directions for future work include investigation of the theoretical properties of \VINE.
\section{Conclusion}
In this work, we have presented \VINE, a non-parametric D-vine copula-based MI estimator. After deriving an analytical expression for the true MI under the simplified D-vine assumption, we motivate a fully non-parametric approach, arguing that it provides high flexibility to capture diverse dependency structures. Through experiments on a variety of benchmarking datasets, we demonstrated that \VINE achieves competitive accuracy to SotA neural methods at a considerably lower computational cost. We also demonstrate that the decompositional structure of \VINE allows for inherent transparency through visualizations of dependencies, which is not offered by black-box neural MI estimators.

\begin{ack}
    The authors would like to thank Arnoldo Frigessi and Nils Lid Hjort for helpful discussions and valuable feedback on this work. 

    This work was supported by the Research Council of Norway through its Centres of Excellence scheme, Integreat - Norwegian Centre for knowledge-driven machine learning under project number 332645 and FRIPRO Grant under project number 356103.

    The computations were performed on resources provided by the Educloud Research infrastructure at the University of Oslo. 
\end{ack}

\newpage
\bibliographystyle{unsrtnat}
\bibliography{refs}

\newpage
\onecolumn
\makeatletter
\let\addcontentsline\origaddcontentsline
\makeatother

\etocsettocstyle{\section*{Table of Contents}}{}
\etocsetnexttocdepth{subsection}
\etoctableofcontents

\newpage
\crefalias{section}{appendix}
\crefalias{subsection}{appendix}
\appendix
\section{Notation and full preliminaries}\label{app:prem}

\paragraph{Notation.}~Throughout this work, we use the lowercase letter $f$ to denote probability density functions (PDFs) and the uppercase letter $F$ to denote cumulative distribution functions (CDFs). Subscripts specify associated variables the functions refer to. Capital letters, such as $X_l, Y_k$, denote univariate random variables. Bold capital letters, such as $\boldsymbol{X}, \boldsymbol{Y}$ denote random vectors, while bold and lowercase letters, such as $\boldsymbol{x}, \boldsymbol{y}$, denote deterministic vectors.

\subsection{Additional details on vine copulas}
\label{app:vines}
This summary is based on \cite{czado_analyzing_2019}.
Let $U_1, \dots, U_d \sim \text{Uniform}(0,1)$. We define the copula $C$ as their joint CDF,
\begin{align*}
    C(u_1, \dots, u_d) = \Pr(U_1 \leq u_1, \dots, U_d \leq u_d).
\end{align*}
For absolutely continuous variables, we obtain the copula density by partial differentiation, $c(u_1, \dots, u_d) = \partial^d C(u_1, \dots, u_d) / (\partial u_1 \cdots \partial u_d)$. For bivariate random variables, there are many such copula functions \cite[Chapter~3]{czado_analyzing_2019}. The simplest being the independence copula, which corresponds to $c(F_1(Z_1),F_2(Z_2)) = 1$, i.e., $Z_1 \perp Z_2$. Other popular choices include the Gaussian copula, which models the dependence between $\Phi^{-1}(F_1(Z_1))$ and $\Phi^{-1}(F_2(Z_2))$ as linear, where $\Phi$ is the standard Gaussian CDF. We also have Archimedean copulas, among those the Clayton copula, which can capture tail dependence. In higher dimensions, one can construct highly flexible dependence structures by combining such bivariate building blocks.

Vine copulas are a class of pair copula constructions, which are constructed by first decomposing the joint PDF into a product of univariate conditional PDFs, 
\begin{align*}
    f(z_1, \dots, z_d) = f_1(z_1) \prod_{j=2}^d f_{j | 1, \dots, j-1}(z_{j} | z_1, \dots, z_{j-1}).
\end{align*}
The idea is that we can express each of these univariate conditional PDFs in terms of bivariate copulas, i.e., pair copulas. We demonstrate this with a 3-dimensional example. The joint PDF $f$ of $(Z_1,Z_2,Z_3)$ can be decomposed as follows\footnote{Note that this decomposition is not unique, and that we in general have $d!$ possible decompositions for a $d$-dimensional PDF.}
\begin{align*}
    f(z_1,z_2,z_3) = f_1(z_1) f_{2|1}(z_2|z_1) f_{3|1,2}(z_3 | z_1, z_2).
\end{align*}
For the first term, we get
\begin{align*}
    f_{2|1}(z_2|z_1) &= \frac{f_{12}(z_1,z_2)}{f_1(z_1)} \\
    &= \frac{f_1(z_1) f_2(z_2) c_{12}(F_1(z_1), F_2(z_2)}{f_1(z_1)} \\
    &= f_2(z_2) c_{12}(F_1(z_1), F_2(z_2).
\end{align*}
Further, for the second term, we get 
\begin{align*}
    f_{3|1,2}(z_3 | z_1, z_2) &= \frac{f_{1,3|2}(z_2, z_3|z_1)}{f_{1|2}(z_2|z_1)} \\
    &= \frac{f_{1|2}(z_1|z_2) f_{3|2}(z_2|z_2) c_{1,3;2}(F_{1|2}(z_1|z_2), F_{3|2}(z_3|z_2))}{f_{1|2}(z_1 | z_2)} \\
    &= f_{3|2}(z_2|z_2) c_{1,3;2}(F_{1|2}(z_1|z_2), F_{3|2}(z_3|z_2)) \\
    &= f_{3}(z_3) c_{2,3}(F_2(z_2), F_3(z_3)) c_{1,3;2}(F_{1|2}(z_1|z_2), F_{3|2}(z_3|z_2)).
\end{align*}
Combining all terms gives us the pair copula decomposition
\begin{align*}
    f(z_1,z_2,z_3) &= f_1(z_1) \times f_2(z_2) \times f_3(z_3) \\
    &\times c_{1,2}(F_{1}(z_1), F_{2}(z_2)) \times c_{2,3}(F_{2}(z_2), F_{3}(z_3)) \\
    &\times c_{1,3;2}(F_{1|2}(z_1 | z_2), F_{3|2}(z_3 | z_2)).
\end{align*}

A vine copula on $d$ variables is constructed in a similar manner, such that the pair copulas can be organized into a hierarchy of trees $\mathcal{T} = (T_1, \dots, T_{d-1})$, called an \emph{R-vine tree sequence}:
\begin{definition}[R-vine tree sequence]   
    \label{def:rvine-treeseq}
    Let $\mathcal{T}=(T_1, \dots, T_{d-1})$ be a sequence of trees. Then $\mathcal{T}$ is said to be an \emph{R-vine tree sequence} if the following holds:
    \begin{enumerate}
    \item Each tree $T_t$ consists of a node set $N_t$ and an edge set $E_t$, $t = 1, \dots, d-1$.
    \item The edges in tree $t$ becomes nodes in tree $t+1$, i.e., $N_t = E_{t-1}$ for $t=2,\dots,d-1$.
    \item An edge in tree $t \geq 2$ must satisfy the proximity condition: two edges in $T_t$ can only form an edge in tree $T_{t+1}$ if they share all variables except for one each.
\end{enumerate}
\end{definition}
In tree $T_t$, an edge $e \in E_t$ represents a pair copula, and in the first trees these model the dependencies between unconditional pairs of variables, while in trees after the first they model dependencies between pairs of variables conditional on others. The proximity condition of \cref{def:rvine-treeseq} ensures that all conditional CDFs, being arguments to pair copulas in trees after the first, can be obtained by partially differentiating pair copulas in the previous trees. In particular, we can use the following result from \citet{joe_families_1996}:
\begin{theorem}[Recursive definition of conditional CDFs]
    Suppose $(Z_1, \dots, Z_d)$ is a random vector, and consider any pair $j,k \in \{ 1,\dots,d \}$ with $j \neq k$, and subset $D \subseteq \{ 1, \dots, d \} \setminus \{ j, k \}$. Then, the following holds:
    \begin{align*}
    F_{j|k, D}(z_{j} | z_{k}, z_{D}) = \frac{\partial C_{j, k; D}(F_{j | D}(z_{j | z_{D}}), F_{k | D}(z_{k | z_{D}})}{\partial F_{k | D}(z_{k} | z_{D})}.
\end{align*}
\end{theorem}
This property is utilized in the sequential estimation and selection algorithm for parametric vines by \citet{dismann_statistical_2010}. That is, the trees are estimated top-down, and estimates from earlier trees are used to generate pseudo-observations for the estimation of the next tree. The same principle also applies to the non-parametric estimator proposed by \citet{nagler_evading_2016}, except that the estimated conditional CDFs are computed from the estimated non-parametric pair copulas.

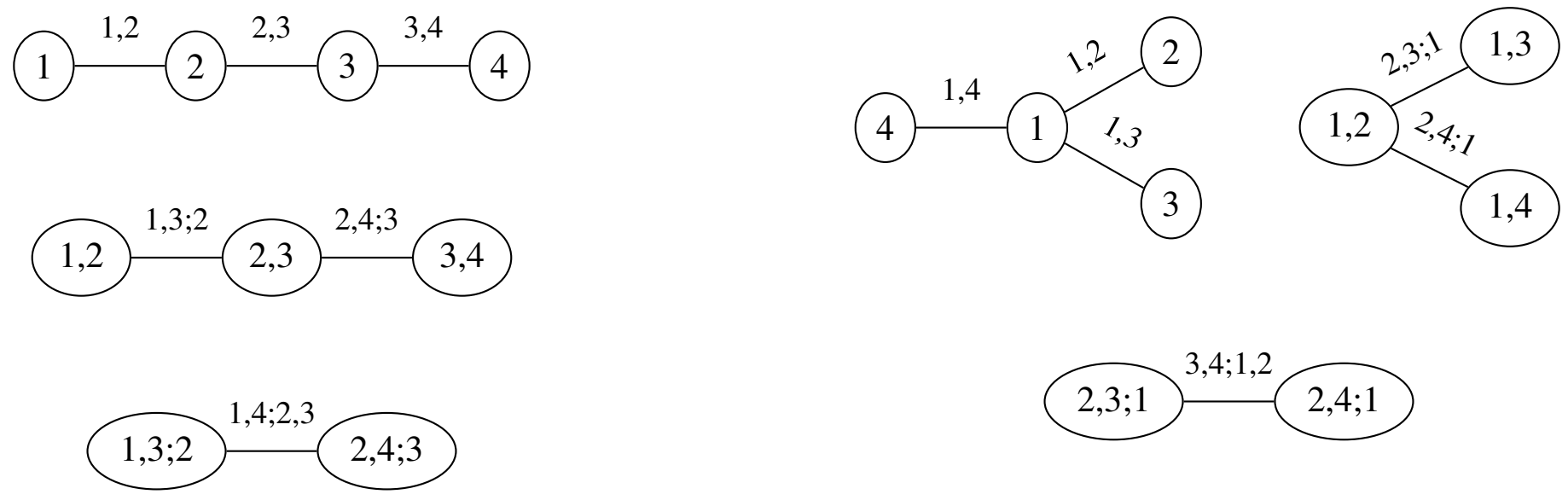
\begin{figure}[h]
\centering

\begin{subfigure}{0.49\textwidth}
\centering

\threefiguresstacked
{
\begin{tikzpicture}
    \begin{scope}[nodes=V]
        \node (1) {1};
        \node (2) [right = 7mm of 1] {2};
        \node (3) [right = 7mm of 2] {3};
        \node (4) [right = 7mm of 3] {4};
    \end{scope}
            
    \draw (1) edge["{1,2}"] (2);
    \draw (2) edge["{2,3}"] (3);
    \draw (3) edge["{3,4}"] (4);
\end{tikzpicture}
}
{
\begin{tikzpicture}
    \begin{scope}[nodes=V]
        \node (1) {1,2};
        \node (2) [right = 7mm of 1] {2,3};
        \node (3) [right = 7mm of 2] {3,4};
    \end{scope}
            
    \draw (1) edge["{1,3;2}"] (2);
    \draw (2) edge["{2,4;3}"] (3);
\end{tikzpicture}
}
{
\begin{tikzpicture}
    \begin{scope}[nodes=V]
        \node (1) {1,3;2};
        \node (2) [right = 7mm of 1] {2,4;3};
    \end{scope}
            
    \draw   (1) edge["{1,4;2,3}"] (2);
\end{tikzpicture}
}

\caption{D-vine tree sequence.}
\label{fig:dvine}
\end{subfigure}
\hfill
\begin{subfigure}{0.49\textwidth}
\centering

\threefiguresgrid
{
\begin{tikzpicture}
    \begin{scope}[nodes=V]
        \node (1) {1};
        \node (2) [above right = 2mm and 7mm of 1] {2};
        \node (3) [below right = 2mm and 7mm of 1] {3};
        \node (4) [left = 7mm of 1] {4};
    \end{scope}
          
    \draw   (1) edge["{1,2}"] (2)
            (1) edge["{1,3}"] (3)
            (1) edge["{1,4}"] (4);
\end{tikzpicture}
}
{
\begin{tikzpicture}
    \begin{scope}[nodes=V]
        \node (1) {1,2};
        \node (2) [above right = 2mm and 7mm of 1] {1,3};
        \node (3) [below right = 2mm and 7mm of 1] {1,4};
    \end{scope}
          
    \draw   (1) edge["{2,3;1}"] (2)
            (1) edge["{2,4;1}"] (3);
\end{tikzpicture}
}
{
\begin{tikzpicture}
    \begin{scope}[nodes=V]
        \node (1) {2,3;1};
        \node (2) [right = 7mm of 1] {2,4;1};
    \end{scope}
          
    \draw   (1) edge["{3,4;1,2}"] (2);
\end{tikzpicture}
}
\caption{C-vine tree sequence.}
\label{fig:cvine}
\end{subfigure}

\caption{Examples of R-vine tree sequences on $d=4$ variables.}
\label{fig:rvine-tree-sequences}
\end{figure}

There are three classes of vine copulas that satisfy \cref{def:rvine-treeseq}. We have visualized two of these classes for $d=4$ variables in \cref{fig:rvine-tree-sequences}. \cref{fig:dvine} shows an example of a D-vine tree sequence. These are characterized by trees where each node is connected to at most two other nodes. This is also the vine tree sequence used in our proposed method. \cref{fig:cvine} shows an example of a C-vine. In a C-vine, each tree has a root note, i.e., a node that is fully connected. The last class of vine copulas is the general R-vines, encompassing all vine copulas, including D-vines and C-vines.

For a D-vine on $d$ variables, the full joint PDF can be written as follows:
\begin{align*}
    f &= \prod_{l=1} f_l \times \prod_{j=1}^{d-1} \prod_{k=1}^{d-j} c_{k, j+k; k+1, \dots, j+k-1}(F_{k|k+1, \dots, j+k-1}, F_{j+k|k+1, \dots, j+k-1}),
\end{align*}
where we have omitted the function arguments of the marginal PDFs and the CDFs for notational convenience.

\paragraph{The simplifying assumption}
In \cref{eq:vinecop}, we see that the pair copula $c_{j_e,k_e;D_e}$ depends on the conditioning value $\boldsymbol{z}_{D_e}$ through its arguments, but we have omitted the direct dependence on the function itself. Generally, it holds that $F_{j_e|D_e}(Z_{j_e}|\boldsymbol{Z}_{D_e}) \perp \boldsymbol{Z}_{D_e}$ and $F_{k_e|D_e}(Z_{k_e}|\boldsymbol{Z}_{D_e}) \perp \boldsymbol{Z}_{D_e}$. This means that all marginal effects of $Z_{j_e} | \boldsymbol{Z}_{D_e} = \boldsymbol{z}_{D_e}$ and $Z_{k_e} | \boldsymbol{Z}_{D_e} = \boldsymbol{z}_{D_e}$ can be captured through the conditional CDFs $F_{j_e|D_e}$ and $F_{k_e|D_e}$, respectively, for all values of $\boldsymbol{z}_{D_e}$. On the other hand, it does not generally hold that $(F_{j_e | D_e}(Z_{j_e} | \boldsymbol{Z}_{D_e}), F_{k_e | D_e}(Z_{k_e} | \boldsymbol{Z}_{D_e})) \perp \boldsymbol{Z}_{D_e}$, meaning that the dependence between $Z_{j_e}$ and $Z_{k_e}$ given $\boldsymbol{Z}_{D_e} = \boldsymbol{z}_{D_e}$ can vary with the conditioning value $\boldsymbol{z}_{D_e}$, in which case $c_{j_e,k_e;D_e}$ would also depend on $\boldsymbol{z}_{D_e}$. An important assumption for making the estimation of vine copulas tractable in higher dimensions is that $c_{j_e, jk_e; D_e}(\cdot,\cdot; \boldsymbol{z}_{D_e}) = c_{j_e, k_e; D_e}(\cdot,\cdot)$ for all possible values of $\boldsymbol{z}_{D_e}$ and all $e \in E_k, k=2,\dots,d-1$. This assumption is known in the vine copula literature as the \emph{simplifying assumption}. Discussions of this assumption can be found in \cite{hobaek_haff_simplified_2010, stober_simplified_2013, nagler_simplified_2025}. While there are distributions that can be represented by simplified vines, such as the multivariate Gaussian and t distributions \cite{stober_simplified_2013}, the simplifying assumption is generally an approximation, a limitation one must be aware of when using such models. 

\subsection{MI estimation} \label{app:mi_est}
MI is a dependency measure able to capture non-linear relationships between random variables. Since it is a function of the joint distribution, it can be estimated from a finite dataset $\{ (\boldsymbol{x}^{(i)}, \boldsymbol{y}^{(i)}) \}_{i=1}^n$.

The earliest approaches to estimating MI were through direct estimation of the joint distribution. These approaches include histogram-based binning, KDE \citep{moon1995estimation}, and $k$-NN \citep{kraskov2004estimating}, all of which are non-parametric methods. Although flexible, all these approaches suffer from the curse of dimensionality, where the sample size required to maintain estimation accuracy grows exponentially with the dimension \citep{mcallester2020formal}. 

To beat the curse of dimensionality faced by non-parametric estimators, recent research has seen the rise of neural MI estimators. Two main strategies are density estimation via deep generative models and density ratio estimation. Deep generative models, such as normalizing flows and variational autoencoders, attempt to learn an explicit parameterization of the underlying densities \citep{cheng2020club, franzese2024minde, mcallester2020formal, song2019understanding}. The density ratio estimation approach trains a neural net to separate samples from the joint distribution $f$ against samples from the product of marginals $f_X f_Y$. This framework includes the Mutual Information Neural Estimator (MINE), which optimizes the Donsker-Varadhan lower bound, and InfoNCE, which utilizes a contrastive objective where the estimate is notably bounded by the logarithm of the batch size \citep{oord2018representation, belghazi2018mutual}. The Smoothing Mutual Information Lower-bound Estimator (SMILE) was proposed to mitigate the high variance often encountered in these bounds by clipping the density ratio \citep{song2019understanding}. Approaches such as MIENF simplify high-dimensional distributions before applying classical estimators such as KDE \citep{butakov2024mutual}. 

An alternative perspective involves isolating the dependency structure from the marginals using copula theory. \citet{safaai2018information} utilized this for a non-parametric copula-based estimator using direct kernel estimation, but only considered the 2-dimensional setting. More recently, the Vine Copula Estimator (VCE) framework \citep{chen_neural_2025} extended this concept to high-dimensional vectors by assuming a simplified vector copula structure, where the marginals are learned using normalizing flows and the vector copula assumes a mixture of Gaussian structure.

\subsection{MI estimation with copulas} \label{app:mi_estimation}
Generally, the MI of random vectors ${\boldsymbol{X} \in \mathbb{R}^{d_X}}$ and ${\boldsymbol{Y} \in \mathbb{R}^{d_Y}}$ can be expressed in terms of an interdependence copula density $c$ and intradependence copula densities $c_X$ and $c_Y$. Let $f$ be the joint PDF of $(\boldsymbol{X},\boldsymbol{Y})$, and let $f_{1}, \dots, f_{d_X}$ denote the marginal PDFs of $X_1, \dots, X_{d_X}$, and $f_{d_X+1}, \dots, f_{d}$ the marginal PDFs of $Y_1, \dots, Y_{d_Y}$. Then, by \cref{thm:sklar} we know that ${f(\boldsymbol{x},\boldsymbol{y}) = \prod_{l=1}^{d_X} f_l(x_l) \times \prod_{l=1}^{d_Y} f_{d_X+l}(y_l) \times c(\boldsymbol{u}_X, \boldsymbol{u}_Y)}$, ${f(\boldsymbol{x}) = \prod_{l=1}^{d_X} f_{l}(x_l) \times c_X(\boldsymbol{u}_x)}$, and ${f_Y(\boldsymbol{y}) = \prod_{l=1}^{d_Y} f_{d_X+l}(y_l) \times c_Y(\boldsymbol{u}_Y)}$, where we have let $\boldsymbol{u}_X = (F_{1}(x_1), \dots, F_{{d_X}}(x_{d_X}))$ and $\boldsymbol{u}_{Y} = (F_{d_X+1}(y_1), \dots, F_{d}(y_{d_Y}))$. Thus, the density ratio can be expressed as
\begin{align*}
    \frac{f(\boldsymbol{x},\boldsymbol{y})}{f_X(\boldsymbol{x}) f_Y(\boldsymbol{y})} = \frac{c(\boldsymbol{u}_X, \boldsymbol{u}_Y)}{c_X(\boldsymbol{u}_X)c_Y(\boldsymbol{u}_Y)}.
\end{align*}
In \cref{app:proofs} we prove \cref{prop:mi-for-dvine}, which states how this expression simplifies under a simplified D-vine copula assumption.

\section{Proof of \cref{prop:mi-for-dvine}}
\label{app:proofs}
\begin{proof}
To derive the MI of a D-vine, we need the joint copula density $c$ of 
$(\boldsymbol{X},\boldsymbol{Y})$, as well as the marginal copula densities $c_{X}$ and 
$c_{Y}$ of $\boldsymbol{X}$ and $\boldsymbol{Y}$, respectively. We have
\begin{align*}
& c(F_{1}(x_{1}),\ldots,F_{d_{X}}(x_{d_{X}}),F_{d_{X}+1}(y_{1}),\ldots,F_{d}(y_{d_{Y}}))\\ 
= & c(F_{1}(z_{1}),\ldots,F_{d_{X}}(z_{d_{X}}),F_{d_{X}+1}(z_{d_{X}+1}),\ldots,F_{d}(z_{d}))\\
= & \prod_{t=1}^{d_{X}-1}\prod_{i=1}^{d_{X}-t}c_{i,i+t;i+1,\ldots,i+t-1}(F_{i|i+1,\ldots,i+t-1}(z_{i}|z_{i+1},\ldots,z_{i+t-1}),\\
& \ \ \ \ \ \ \ \ \ \ \ \ \ \ \ \ \ \ \ \ \ \ \ \ \ \ \ \ \ \ \ \ \ \ \ \ \ \ \ \ \ \ \ \ \ \ \ \ F_{i+t|i+1,\ldots,i+t-1}(z_{i+t}|z_{i+1},\ldots,z_{i+t-1}))\\
\cdot & \prod_{t=1}^{d_{Y}-1}\prod_{i=d_{X}+1}^{d-t}c_{i,i+t;i+1,\ldots,i+t-1}(F_{i|i+1,\ldots,i+t-1}(z_{i}|z_{i+1},\ldots,z_{i+t-1}),\\
& \ \ \ \ \ \ \ \ \ \ \ \ \ \ \ \ \ \ \ \ \ \ \ \ \ \ \ \ \ \ \ \ \ \ \ \ \ \ \ \ \ \ \ \ \ \ \ \ \ \ \ \ F_{i+t|i+1,\ldots,i+t-1}(z_{i+t}|z_{i+1},\ldots,z_{i+t-1}))\\
\cdot & \prod_{t=1}^{\text{min}(d_{X},d_{Y})-1}\prod_{i=d_{X}+1-t}^{d_{X}}c_{i,i+t;i+1,\ldots,i+t-1}(F_{i|i+1,\ldots,i+t-1}(z_{i}|z_{i+1},\ldots,z_{i+t-1}),\\
& \ \ \ \ \ \ \ \ \ \ \ \ \ \ \ \ \ \ \ \ \ \ \ \ \ \ \ \ \ \ \ \ \ \ \ \ \ \ \ \ \ \ \ \ \ \ \ \ \ \ \ \ \ \ \ \ \ \ \ \ \ \ \ \ \ \ \ \ F_{i+t|i+1,\ldots,i+t-1}(z_{i+t}|z_{i+1},\ldots,z_{i+t-1}))\\
\cdot & \prod_{t=\text{min}(d_{X},d_{Y})}^{\text{max}(d_{X},d_{Y})-1}\prod_{i=1+\text{max}(d_{X}-t,0)}^{\text{min}(d_{X},d_{Y})+\text{max}(d_{X}-t,0)}c_{i,i+t;i+1,\ldots,i+t-1}(F_{i|i+1,\ldots,i+t-1}(z_{i}|z_{i+1},\ldots,z_{i+t-1}),\\
& \ \ \ \ \ \ \ \ \ \ \ \ \ \ \ \ \ \ \ \ \ \ \ \ \ \ \ \ \ \ \ \ \ \ \ \ \ \ \ \ \ \ \ \ \ \ \ \ \ \ \ \ \ \ \ \ \ \ \ \ \ \ \ \ \ \ \ \ \ \ \ \ \ \ \ \ \ \ \ \ \ \ \ \ \ \ \ \ \ F_{i+t|i+1,\ldots,i+t-1}(z_{i+t}|z_{i+1},\ldots,z_{i+t-1}))\\
\cdot & \prod_{t=\text{max}(d_{X},d_{Y})}^{d-1}\prod_{i=1}^{d-t}c_{i,i+t;i+1,\ldots,i+t-1}(F_{i|i+1,\ldots,i+t-1}(z_{i}|z_{i+1},\ldots,z_{i+t-1}),\\
& \ \ \ \ \ \ \ \ \ \ \ \ \ \ \ \ \ \ \ \ \ \ \ \ \ \ \ \ \ \ \ \ \ \ \ \ \ \ \ \ \ \ \ \ \ \ \ \ \ \ \ \ \ \ \ \ \ \ F_{i+t|i+1,\ldots,i+t-1}(z_{i+t}|z_{i+1},\ldots,z_{i+t-1})),
\end{align*}
where the first two double products correspond to the orange and blue 
area, respectively, of \cref{fig:dvine-decomp}, and the last three 
double products correspond to the green area, given by the edges in 
$\E_{XY}$. Grouping the last four double products, we may re-write 
the joint copula density as
\begin{align*}
& c(F_{1}(z_{1}),\ldots,F_{d_{X}}(z_{d_{X}}),F_{d_{X}+1}(z_{d_{X}+1}),\ldots,F_{d}(z_{d}))\\
= & \prod_{t=1}^{d_{X}-1}\prod_{i=1}^{d_{X}-t}c_{i,i+t;i+1,\ldots,i+t-1}(F_{i|i+1,\ldots,i+t-1}(z_{i}|z_{i+1},\ldots,z_{i+t-1}),\\
& \ \ \ \ \ \ \ \ \ \ \ \ \ \ \ \ \ \ \ \ \ \ \ \ \ \ \ \ \ \ \ \ \ \ \ \ \ \ \ \ \ \ \ \ \ \ \ \ F_{i+t|i+1,\ldots,i+t-1}(z_{i+t}|z_{i+1},\ldots,z_{i+t-1}))\\
\cdot & \prod_{j=d_{X}+1}^{d}\prod_{l=1}^{j-2}c_{l,j;l+1,\ldots,j-1}(F_{l|l+1,\ldots,j-1}(z_{l}|z_{l+1},\ldots,z_{j-1}),\\
& \ \ \ \ \ \ \ \ \ \ \ \ \ \ \ \ \ \ \ \ \ \ \ \ \ \ \ \ \ \ \ \ \ \ \ \ \ \ \ \ \ \ F_{j|l+1,\ldots,j-1}(z_{j}|z_{l+1},\ldots,z_{j-1}))\cdot c_{j-1,j}(F_{j-1}(z_{j-1},F_{j}(z_{j}))).
\end{align*}
Using Sklar's theorem, we have
\[
c_{j-1,j}(F_{j-1}(z_{j-1}),F_{j}(z_{j}))) = \frac{f_{j-1,j}(z_{j-1},z_{j})}{f_{j-1}(z_{j-1})f_{j}(z_{j})} = \frac{f_{j|j-1}(z_{j}|z_{j-1})}{f_{j}(z_{j})}
\]
\begin{align*}
&c_{l,j;l+1,\ldots,j-1}(F_{l|l+1,\ldots,j-1}(z_{l}|z_{l+1},\ldots,z_{j-1}),F_{j|l+1,\ldots,j-1}(z_{j}|z_{l+1},\ldots,z_{j-1}))\\
& = \frac{f_{l,j|l+1,\ldots,j-1}(z_{l},z_{j}|z_{l+1},\ldots,z_{j-1})}{f_{l|l+1,\ldots,j-1}(z_{l}|z_{l+1},\ldots,z_{j-1})f_{j|l+1,\ldots,j-1}(z_{j}|z_{l+1},\ldots,z_{j-1})}\\
& = \frac{f_{j|l,\ldots,j-1}(z_{j}|z_{l},\ldots,z_{j-1})}{f_{j|l+1,\ldots,j-1}(z_{j}|z_{l+1},\ldots,z_{j-1})}
\end{align*}
Omitting arguments for simplicity, we obtain
\begin{align*}
\prod_{l=1}^{j-2}c_{l,j;l+1,\ldots,j-1}\cdot c_{j-1,j} = \frac{f_{j|1,\ldots,j-1}}{f_{j|2,\ldots,j-1}}\cdot \frac{f_{j|2,\ldots,j-1}}{f_{j|3,\ldots,j-1}} \cdot \cdots \cdot \frac{f_{j|j-2,j-1}}{f_{j|j-1}} \cdot \frac{f_{j|j-1}}{f_{j}} = \frac{f_{j|1,\ldots,j-1}}{f_{j}}.
\end{align*}
Plugging this into the expression for the joint D-vine copula density, we get
\begin{align*}
& c(F_{1}(z_{1}),\ldots,F_{d_{X}}(z_{d_{X}}),F_{d_{X}+1}(z_{d_{X}+1}),\ldots,F_{d}(z_{d}))\\
= & \prod_{t=1}^{d_{X}-1}\prod_{i=1}^{d_{X}-t}c_{i,i+t;i+1,\ldots,i+t-1}(F_{i|i+1,\ldots,i+t-1}(z_{i}|z_{i+1},\ldots,z_{i+t-1}),\\
& \ \ \ \ \ \ \ \ \ \ \ \ \ \ \ \ \ \ \ \ \ \ \ \ \ \ \ \ \ \ \ \ \ \ \ \ \ \ \ \ \ \ \ \ \ \ \ \ F_{i+t|i+1,\ldots,i+t-1}(z_{i+t}|z_{i+1},\ldots,z_{i+t-1}))\\
\cdot & \prod_{j=d_{X}+1}^{d}\frac{f_{j|1,\ldots,j-1}(z_{j}|z_{1},\ldots,z_{j-1})}{f_{j}(z_{j})}\\
= & \prod_{t=1}^{d_{X}-1}\prod_{i=1}^{d_{X}-t}c_{i,i+t;i+1,\ldots,i+t-1}(F_{i|i+1,\ldots,i+t-1}(z_{i}|z_{i+1},\ldots,z_{i+t-1}),\\
& \ \ \ \ \ \ \ \ \ \ \ \ \ \ \ \ \ \ \ \ \ \ \ \ \ \ \ \ \ \ \ \ \ \ \ \ \ \ \ \ \ \ \ \ \ \ \ \ F_{i+t|i+1,\ldots,i+t-1}(z_{i+t}|z_{i+1},\ldots,z_{i+t-1}))\\
\cdot & \frac{f_{d_{X}+1,\ldots,d|1,\ldots,d_{X}}(z_{d_{X}+1,\ldots,z_{d}}|z_{1},\ldots,z_{d_{X}})}{\prod_{j=d_{X}+1}^{d}f_{j}(z_{j})}.
\end{align*}
The marginal copula density of $\boldsymbol{X}$ is then given by
\begin{align*}
&c_{X}(\boldsymbol{u}_{X})\\
= &\int c(\boldsymbol{u}_{X},\boldsymbol{u}_{Y}) d \boldsymbol{u}_{Y}\\
= &\int c(F_{1}(z_{1}),\ldots,F_{d_{X}}(z_{d_{X}}),F_{d_{X}+1}(z_{d_{X}+1}),\ldots,F_{d}(z_{d})) \cdot \prod_{j=d_{X}+1}^{d}f_{j}(z_{j}) dz_{d_{x}+1}\ldots dz_{d}\\
= &\prod_{t=1}^{d_{X}-1}\prod_{i=1}^{d_{X}-t}c_{i,i+t;i+1,\ldots,i+t-1}(F_{i|i+1,\ldots,i+t-1}(z_{i}|z_{i+1},\ldots,z_{i+t-1}),\\
& \ \ \ \ \ \ \ \ \ \ \ \ \ \ \ \ \ \ \ \ \ \ \ \ \ \ \ \ \ \ \ \ \ \ \ \ \ \ \ \ \ \ \ \ \ \ \ \ F_{i+t|i+1,\ldots,i+t-1}(z_{i+t}|z_{i+1},\ldots,z_{i+t-1}))\\
&\cdot \int f_{d_{X}+1,\ldots,d|1,\ldots,d_{X}}(z_{d_{X}+1,\ldots,z_{d}}|z_{1},\ldots,z_{d_{X}})dz_{d_{x}+1}\ldots dz_{d}\\
& = \prod_{t=1}^{d_{X}-1}\prod_{i=1}^{d_{X}-t}c_{i,i+t;i+1,\ldots,i+t-1}(F_{i|i+1,\ldots,i+t-1}(z_{i}|z_{i+1},\ldots,z_{i+t-1}),\\
& \ \ \ \ \ \ \ \ \ \ \ \ \ \ \ \ \ \ \ \ \ \ \ \ \ \ \ \ \ \ \ \ \ \ \ \ \ \ \ \ \ \ \ \ \ \ \ \ \ \ \ \ F_{i+t|i+1,\ldots,i+t-1}(z_{i+t}|z_{i+1},\ldots,z_{i+t-1})).
\end{align*}
Similarly, we get the marginal copula density of $\boldsymbol{Y}$ 
\begin{align*}
c_{Y}(\boldsymbol{u}_{Y}) = &\prod_{t=d_{X}+1}^{d-1}\prod_{i=1}^{d-t}c_{i,i+t;i+1,\ldots,i+t-1}(F_{i|i+1,\ldots,i+t-1}(z_{i}|z_{i+1},\ldots,z_{i+t-1}),\\
& \ \ \ \ \ \ \ \ \ \ \ \ \ \ \ \ \ \ \ \ \ \ \ \ \ \ \ \ \ \ \ \ \ \ \ \ \ \ \ \ \ \ \ \ \ \ \ \ F_{i+t|i+1,\ldots,i+t-1}(z_{i+t}|z_{i+1},\ldots,z_{i+t-1})).
\end{align*}
Hence, we get
\begin{align*}
&I(\boldsymbol{X};\boldsymbol{Y}) = \int \log\frac{c(\boldsymbol{u}_{X},\boldsymbol{u}_{Y})}{c_{X}(\boldsymbol{u}_{X})c_{Y}(\boldsymbol{u}_{Y})}c(\boldsymbol{u}_{X},\boldsymbol{u}_{Y})d \boldsymbol{u}_{X}d \boldsymbol{u}_{Y}\\
= &\int \log\frac{c(F_{1}(z_{1}),\ldots,F_{d}(z_{d}))}{c_{X}(F_{1}(z_{1}),\ldots,F_{d_{X}}(z_{d_{X}}))c_{Y}(F_{d_{X}+1}(z_{d_{X}+1}),\ldots,F_{d}(z_{d}))}f(z_{1},\ldots,z_{d})dz_{1}\ldots dz_{d}\\
= &\E_{f}\!\left(\sum_{t=1}^{\text{min}(d_{X},d_{Y})-1}\sum_{i=d_{X}+1-t}^{d_{X}}\log c_{i,i+t;i+1,\ldots,i+t-1}(F_{i|i+1,\ldots,i+t-1}(Z_{i}|Z_{i+1},\ldots,Z_{i+t-1}),\right.\\
& \ \ \ \ \ \ \ \ \ \ \ \ \ \ \ \ \ \ \ \ \ \ \ \ \ \ \ \ \ \ \ \ \ \ \ \ \ \ \ \ \ \ \ \ \ \ \ \ \ \ \ \ \ \ \ \ \ \ \ \ \ \ \ \ \ \ \ \ \ \ \ \ \ \ \ \ \ \ \ \ \ F_{i+t|i+1,\ldots,i+t-1}(Z_{i+t}|Z_{i+1},\ldots,Z_{i+t-1}))\\
& \ \ \ \ \ \ \ \ + \sum_{t=\text{min}(d_{X},d_{Y})}^{\text{max}(d_{X},d_{Y})-1}\bigg(\\
& \ \ \ \ \ \ \ \ \sum_{i=1+\text{max}(d_{X}-t,0)}^{\text{min}(d_{X},d_{Y})+\text{max}(d_{X}-t,0)}\log c_{i,i+t;i+1,\ldots,i+t-1}(F_{i|i+1,\ldots,i+t-1}(Z_{i}|Z_{i+1},\ldots,Z_{i+t-1}),\\
& \ \ \ \ \ \ \ \ \ \ \ \ \ \ \ \ \ \ \ \ \ \ \ \ \ \ \ \ \ \ \ \ \ \ \ \ \ \ \ \ \ \ \ \ \ \ \ \ \ \ \ \ \ \ \ \ \ \ \ \ \ \ \ \ \ \ \ \ \ \ \ \ \ \ \ \ \ \ \ \ \ F_{i+t|i+1,\ldots,i+t-1}(Z_{i+t}|Z_{i+1},\ldots,Z_{i+t-1}))\bigg)\\
& \ \ \ \ \ \ \ + \sum_{t=\text{max}(d_{X},d_{Y})}^{d-1}\sum_{i=1}^{d-t}\log c_{i,i+t;i+1,\ldots,i+t-1}(F_{i|i+1,\ldots,i+t-1}(Z_{i}|Z_{i+1},\ldots,Z_{i+t-1}),\\
& \ \ \ \ \ \ \ \ \ \ \ \ \ \ \ \ \ \ \ \ \ \ \ \ \ \ \ \ \ \ \ \ \ \ \ \ \ \ \ \ \ \ \ \ \ \ \ \ \ \ \ \ \ \ \ \ \ \ \ \ \ \ \ \ \ \ \ \ \ \ \ \ \ \ F_{i+t|i+1,\ldots,i+t-1}(Z_{i+t}|Z_{i+1},\ldots,Z_{i+t-1}))\\
= &\E_{f}\!\left(\sum_{e \in E_{XY}} \log c_{j_e, k_e; D_e} \! \left( F_{j_e | D_e}(Z_{j_e} | \boldsymbol{Z}_{D_e}), F_{k_e | D_e}(Z_{k_e} | \boldsymbol{Z}_{D_e})\right)\right).
\end{align*}
\end{proof}

\section{Computational and Scaling Limitations of \VINE}\label{app:limitations}
While \VINE scales to much higher dimensions than traditional non-parametric MI estimators, its practical implementation is subject to computational bottlenecks and numerical instabilities as $d$ grows exceptionally large. Because a D-vine construction requires estimating up to $d(d-1)/2$ non-parametric pair-copulas which cannot be estimated all in parallel, the computational complexity increases greatly with $d$. For standard tabular tasks or moderately-sized latent spaces ($d \le 64$), this overhead is well-managed and often computes faster than the training loops required for neural estimators. However, for heavily uncompressed spaces where $d > 100$, the sheer volume of sequential kernel density estimations becomes a primary computational bottleneck. 

\begin{figure}[H]
    \centering
    \includegraphics[width=0.5\textwidth]{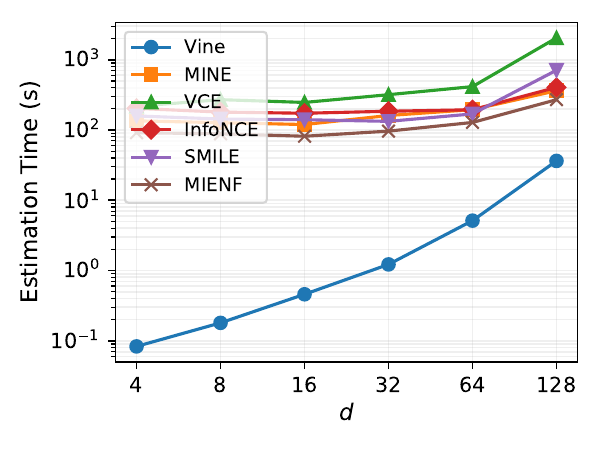}
    \caption{Empirical scaling limitations of the \VINE estimator on a multivariate Gaussian task. As the $d$ increases, the computational time scales quadratically (left axis). For $d > 64$, error accumulation from deep conditional evaluations begins to notably degrade the overall MI estimation accuracy (right axis), illustrating the practical limits of the D-vine decomposition in extremely high-dimensional spaces.}
    \label{fig:scaling_limitations}
\end{figure}

Furthermore, while the theoretical error bound isolating $d^2$ is a polynomial scaling factor rather than an exponential one, numerical error accumulation remains a practical limitation. Because pseudo-observations are generated sequentially through the vine trees, small estimation errors in the conditional CDFs of the initial trees propagate into the inputs of deeper trees. Even with the threshold parameter $\tau$ actively pruning weak dependencies to combat this accumulation, the compounded numerical drift can eventually degrade the accuracy of the estimator when $d$ scales into the hundreds. We illustrate this breakdown point in \cref{fig:scaling_limitations}, which plots both the average computational time and the corresponding loss in estimation accuracy as a function of the dimension. In our empirical evaluations, \VINE successfully maintained stable estimates for dimensions up to $d = 256$; however, when evaluated at $d = 512$, the accumulated approximation errors and numerical instabilities cause the estimator to break down significantly. 

\section{Regularization via the threshold parameter}\label{app:threshold_param}
To evaluate the sensitivity of our estimator to the crucial threshold hyperparameter, we investigate four different datasets with different dimensions. As described in \cref{sec:methods}, the threshold is a pruning mechanism that replaces any bivariate pair-copula with an independence copula if its empirical Kendall's $\tau$ falls below the specified value. This effectively removes weak dependencies in the higher-order trees of the vine, reducing the risk of error accumulation and overfitting. We evaluated the threshold's impact across four diverse tasks: Gaussian, cube-transformed, spiral-mapped, and Student-$t$ distribution. As illustrated in \cref{fig:scaling_limitations}, the estimator is fairly stable and aligned with the ground truth across all values between $10^{-4}$ and $10^{-1}$. For values larger than $10^{-1}$, the estimator begins to prune legitimate dependencies, leading to a significant underestimation of the ground truth MI. 

For every experiment, setting $\tau=0.1$ leads to the most accurate MI estimate. The selection of $\tau$ is particularly critical for the Student-$t$ data. Due to the heavy-tailed nature of the distribution and the high dimensionality of the latent space, non-parametric estimators are susceptible to fitting spurious correlations in the deeper layers of the vine construction. In these regimes, the threshold acts as a vital structural regularizer. By setting a default value of $\tau=0.1$, we effectively filter out sampling noise without requiring ad hoc tuning for specific datasets, something that would be impossible in most realistic scenarios. 
\begin{figure}[tb]
    \centering
    \begin{subfigure}[b]{0.45\textwidth}
        \centering
        \includegraphics[width=\textwidth]{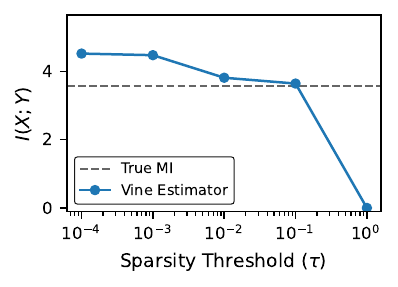}
        \caption{Gaussian Dependency ($d=16$)}
        \label{fig:thresh_gauss}
    \end{subfigure}
    \hfill
    \begin{subfigure}[b]{0.45\textwidth}
        \centering
        \includegraphics[width=\textwidth]{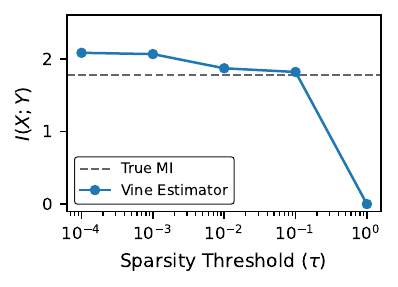}
        \caption{Cube Transformation ($d=8$)}
        \label{fig:thresh_cube}
    \end{subfigure}
    
    \vspace{4mm} 

    \begin{subfigure}[b]{0.45\textwidth}
        \centering
        \includegraphics[width=\textwidth]{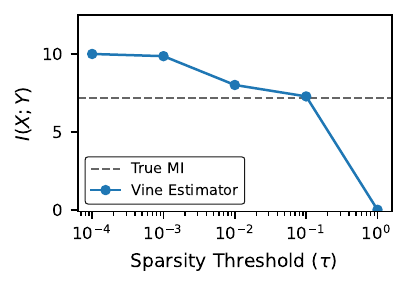}
        \caption{Spiral Mapping ($d=64$)}
        \label{fig:thresh_spiral}
    \end{subfigure}
    \hfill
    \begin{subfigure}[b]{0.45\textwidth}
        \centering
        \includegraphics[width=\textwidth]{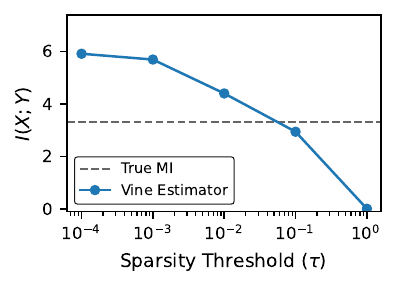}
        \caption{Student-$t$ Distribution ($d=32, \nu=3$)}
        \label{fig:thresh_student}
    \end{subfigure}
    
    \caption{Sensitivity of the Vine Copula Estimator to the sparsity threshold $\tau$ across datasets. The estimator remains mostly stable across three orders of magnitude, with significant information loss occurring only after $\tau > 0.1$. This robustness allows for a universal default threshold without per-task tuning.}
    \label{fig:threshold_sweep}
\end{figure}

\section{Experiment details} \label{app:exp_details}

\subsection{Implementation setup}
\VINE is implemented using the Python package \texttt{pyvinecopulib} \cite{naglerpyvinecopulib2025}. In all experiments, the pair copulas are estimated non-parametrically using Transformation Local-Likelihood (TLL) kernels \cite{geenens_probittransformation_2017} with quadratic estimation method. According to \cref{app:threshold_param}, we set the threshold parameter $\tau = 0.1$. 

The neural benchmarks include MINE, InfoNCE, SMILE, MIENF, VCE, and MINDE. These estimators utilize a standardized neural critic consisting of a three-layer Multi-Layer Perceptron (MLP) architecture with 500 hidden units per layer and ReLU activation functions. All neural models are trained using the Adam optimizer with a learning rate of $5 \times 10^{-4}$, a weight decay of $10^{-5}$, and a batch size of 500. To balance computational overhead with convergence stability, we set the maximum training duration to 500 iterations with an early stopping patience of 200 iterations. For synthetic and transformed Gaussian tasks, the models are trained on 5,000 samples, while the image-based benchmarks utilize 10,000 samples to accommodate the higher-dimensional latent spaces. The vine method models were run on an Apple M4 Max (16-core CPU, 36GB RAM), whereas the neural estimators were run on NVIDIA A1000 GPUs.

\subsection{Transformed Gaussian datasets} \label{app:transformed_gaussian}
This appendix provides the formal definitions for the non-Gaussian data-generating distributions used in our benchmarking suite. Mutual information is invariant under strictly increasing bijective transformations of the margins, but the underlying copula is not generally preserved under such transformations. As a result, applying these transformations offers no guarantee that the simplifying assumption still holds. The only case we consider in which this preservation is guaranteed is when we apply strictly increasing transformations to the margins of multivariate Gaussian distributions. Using these facts, we construct benchmarks with known ground-truth MI, allowing us to investigate \VINE's robustness to violations of model assumptions, particularly the simplifying assumption.

We consider the following mappings applied to multivariate Gaussian distributions:
\begin{itemize}
    \item Strictly increasing transformation of the margins: \textbf{simplifying assumption is preserved}
    \begin{itemize}
        \item {Cube Transformation:} $f(x) = x^3$.
        \item {Hyperbolic Tangent (Tanh):} $f(x) = \tanh(x)$.
        \item {Sign-Square:} $f(x) = \sign(x) \cdot x^2$.
    \end{itemize}
    \item Joint transformation: \textbf{simplifying assumption is not preserved}
    \begin{itemize}
        \item {Spiral:} This task defines a dependency where variables are distributed along a 2D spiral arm. The coordinates are generated via a polar mapping $(x, y) = (\theta \cos \theta, \theta \sin \theta)$, where the angle $\theta$ is coupled between $X$ and $Y$. This creates a relationship that is locally structured but globally complex, requiring the estimator to capture a "winding" dependency that spans multiple rotations of the support.
        \item {Swiss roll:} A classic manifold learning benchmark where a 2D surface is "rolled" into a 3D volume. The dependency is defined on the surface of the roll, which is then embedded into a $d$-dimensional space. This tests whether an estimator can successfully isolate the underlying low-dimensional manifold structure from the high-dimensional ambient noise and redundant features.
    \end{itemize}
\end{itemize}
In addition to transformations of Gaussian distributions, we include the following data-generating distributions in our experiments:
\begin{itemize}
    \item {Multivariate $t$-distribution}: \textbf{simplifying assumption holds}. The true MI can be determined analytically from the distribution parameters.
    \item {Mixture of Gaussian (MoG) distributions}: \textbf{simplifying assumption does not hold}. The true MI can be estimated with high precision via Monte Carlo integration.
\end{itemize}

\subsection{Image dataset} \label{app:image}

\textbf{Experiment details:}
Neural estimators were trained with a three-layer MLP architecture (500 units per layer) for 150 iterations per estimation. Our Vine estimator utilizes non-parametric Transformation Local Likelihood (TLL) kernels for bivariate pair-copula estimation to capture complex, non-linear dependencies without making restrictive parametric assumptions.

\begin{table}[htbp]
\centering
\caption{Cumulative Explained Variance Captured by PCA for Different Latent Dimensions ($d$). Values represent the percentage of total variance preserved for the input ($X$) and target ($Y$) image sets.}
\label{tab:pca_variance}
\begin{tabular}{@{}lccccc@{}}
\toprule
\textbf{Dataset} & \textbf{Dim ($d$)} & \textbf{Var ($X$)} & \textbf{Var ($Y$)} & \textbf{Chosen?} \\ \midrule
\multirow{4}{*}{Gaussian} & 4  & 83.04\% & 82.95\% & No \\
                          & 8  & 92.59\% & 93.58\% & No \\
                          & 12 & 97.27\% & 97.26\% & \textbf{Yes} \\
                          & 16 & 99.86\% & 99.86\% & --- \\ \midrule
\multirow{4}{*}{Rectangle} & 20 & 88.67\% & 88.58\% & No \\
                                   & 30 & 92.75\% & 92.74\% & No \\
                                   & 40 & 95.07\% & 95.09\% & \textbf{Yes} \\
                                   & 50 & 96.63\% & 96.65\% & --- \\ \bottomrule
\end{tabular}
\end{table}

\newpage
\section*{NeurIPS Paper Checklist}

\begin{enumerate}

\item {\bf Claims}
    \item[] Question: Do the main claims made in the abstract and introduction accurately reflect the paper's contributions and scope?
    \item[] Answer: \answerYes{} 
    \item[] Justification: The abstract and introduction clearly outline the paper’s primary contributions: an efficient, interpretable MI estimator (VINE) that leverages vine copula theory to bypass the curse of dimensionality. These claims are supported by theoretical motivations in \cref{sec:guarantees} and empirical results across synthetic and image datasets in Section 5.

\item {\bf Limitations}
    \item[] Question: Does the paper discuss the limitations of the work performed by the authors?
    \item[] Answer: \answerYes{} 
    \item[] Justification: Limitations are explicitly discussed in Section 6.

\item {\bf Theory assumptions and proofs}
    \item[] Question: For each theoretical result, does the paper provide the full set of assumptions and a complete (and correct) proof?
    \item[] Answer: \answerYes{} 
    \item[] Justification: The paper states the assumptions for its theoretical contribution, Proposition 3.1, and provides a proof in Appendix B. Furthermore, it discusses the necessary conditions for the "simplifying assumption" in Section 2.1 and references the specific regularity conditions (A1–A4) required for the convergence rate claims in Section 4.

    \item {\bf Experimental result reproducibility}
    \item[] Question: Does the paper fully disclose all the information needed to reproduce the main experimental results of the paper to the extent that it affects the main claims and/or conclusions of the paper (regardless of whether the code and data are provided or not)?
    \item[] Answer: \answerYes{} 
    \item[] Justification: All experimental setups are explained, and all hyperparameters are given in detail in \cref{app:exp_details}.

\item {\bf Open access to data and code}
    \item[] Question: Does the paper provide open access to the data and code, with sufficient instructions to faithfully reproduce the main experimental results, as described in supplemental material?
    \item[] Answer: \answerYes{} 
    \item[] Justification: We have provided implementation details and hyperparameter settings in the supplemental material to ensure reproducibility.

\item {\bf Experimental setting/details}
    \item[] Question: Does the paper specify all the training and test details (e.g., data splits, hyperparameters, how they were chosen, type of optimizer) necessary to understand the results?
    \item[] Answer: \answerYes{} 
    \item[] Justification: All hyperparameters and experimental details are given in \cref{app:exp_details}.

\item {\bf Experiment statistical significance}
    \item[] Question: Does the paper report error bars suitably and correctly defined or other appropriate information about the statistical significance of the experiments?
    \item[] Answer: \answerYes{} 
    \item[] Justification: For the experiment in Section 5.3, we report standard error for every estimator. However, we do not report standard error for every experiment due to the computational time.

\item {\bf Experiments compute resources}
    \item[] Question: For each experiment, does the paper provide sufficient information on the computer resources (type of compute workers, memory, time of execution) needed to reproduce the experiments?
    \item[] Answer: \answerYes{} 
    \item[] Justification: This is provided in \cref{app:exp_details}.
    
\item {\bf Code of ethics}
    \item[] Question: Does the research conducted in the paper conform, in every respect, with the NeurIPS Code of Ethics \url{https://neurips.cc/public/EthicsGuidelines}?
    \item[] Answer: \answerYes{} 
    \item[] Justification: The research is a foundational study focusing on a new methodology for mutual information estimation using synthetic and publicly available benchmark datasets. It does not involve human subjects, sensitive personal data, or any data collection practices that conflict with the NeurIPS Code of Ethics.

\item {\bf Broader impacts}
    \item[] Question: Does the paper discuss both potential positive societal impacts and negative societal impacts of the work performed?
    \item[] Answer: \answerNA{} 
    \item[] Justification: This work by its nature is focused on mutual information estimation, and has no direct societal impact.
    
\item {\bf Safeguards}
    \item[] Question: Does the paper describe safeguards that have been put in place for responsible release of data or models that have a high risk for misuse (e.g., pre-trained language models, image generators, or scraped datasets)?
    \item[] Answer: \answerNA{} 
    \item[] Justification: This work only uses synthetic data and standard synthetic image benchmarks.The project does not involve the release of pre-trained generative models, large-scale language models, or sensitive scraped datasets that carry a high risk of misuse or require specific safety safeguards.

\item {\bf Licenses for existing assets}
    \item[] Question: Are the creators or original owners of assets (e.g., code, data, models), used in the paper, properly credited and are the license and terms of use explicitly mentioned and properly respected?
    \item[] Answer: \answerYes{} 
    \item[] Justification: We properly credit the original creators of all datasets and baseline mutual information estimators (e.g., VCE, MINE, InfoNCE, and other MI estimators) through formal
citations.

\item {\bf New assets}
    \item[] Question: Are new assets introduced in the paper well documented and is the documentation provided alongside the assets?
    \item[] Answer: \answerNA{} 
    \item[] Justification: No new assets are released in this work.

\item {\bf Crowdsourcing and research with human subjects}
    \item[] Question: For crowdsourcing experiments and research with human subjects, does the paper include the full text of instructions given to participants and screenshots, if applicable, as well as details about compensation (if any)? 
    \item[] Answer: \answerNA{} 
    \item[] Justification: No experiments are performed using human subjects.

\item {\bf Institutional review board (IRB) approvals or equivalent for research with human subjects}
    \item[] Question: Does the paper describe potential risks incurred by study participants, whether such risks were disclosed to the subjects, and whether Institutional Review Board (IRB) approvals (or an equivalent approval/review based on the requirements of your country or institution) were obtained?
    \item[] Answer: \answerNA{} 
    \item[] Justification: No experiments are performed using human subjects.

\item {\bf Declaration of LLM usage}
    \item[] Question: Does the paper describe the usage of LLMs if it is an important, original, or non-standard component of the core methods in this research? Note that if the LLM is used only for writing, editing, or formatting purposes and does \emph{not} impact the core methodology, scientific rigor, or originality of the research, declaration is not required.
    \item[] Answer: \answerNA{} 
    \item[] Justification: Large Language Models were not a component of the core methodology, scientific rigor, or originality of the research.

\end{enumerate}

\end{document}